\documentclass[11pt]{article}

\usepackage[final]{acl}

\usepackage{times}
\usepackage{latexsym}

\usepackage[T1]{fontenc}

\usepackage[utf8]{inputenc}

\usepackage{microtype}

\usepackage{inconsolata}

\usepackage{graphicx}

\usepackage{times,latexsym}
\usepackage[T1]{fontenc}
\usepackage{microtype}

\usepackage{amsmath}
\usepackage{amsthm}
\usepackage{amsfonts}
\usepackage{amssymb}
\usepackage{mathtools}
\usepackage{bm}

\usepackage{tikz}
\usepackage{adjustbox}
\usetikzlibrary{bayesnet}
\usetikzlibrary{arrows}
\usetikzlibrary{decorations}
\usepackage{rotating}

\usepackage{graphicx}
\usepackage{booktabs}
\usepackage{tabularx}
\usepackage{tabulary}
\usepackage{colortbl}
\usepackage{xspace}
\usepackage{xcolor}
\newcolumntype{Y}{>{\centering\arraybackslash}X}
  \newcolumntype{P}{>{\raggedleft\arraybackslash}X}
\usepackage{multirow}
\usepackage{stfloats}
\usepackage{algorithm}
\usepackage[noend]{algorithmic}

\usepackage[noabbrev,capitalise]{cleveref}
\usepackage{hyperref}

\usepackage{arydshln}

\usepackage{makecell}
\usepackage{array}

\usepackage{colortbl}
\usepackage{booktabs}
\usepackage[table]{xcolor}

\definecolor{lvbg}{RGB}{240,248,255} 
\definecolor{vbg}{RGB}{255,245,238}

\usepackage{todonotes}
\usepackage{bm}

\makeatletter
\newcommand*\iftodonotes{\if@todonotes@disabled\expandafter\@secondoftwo\else\expandafter\@firstoftwo\fi}
\makeatother

\definecolor{edolime}{rgb}{0.9,1,0.3}

\newcommand{\uidlv}{\mathrm{UID}_{lv}\xspace}
\newcommand{\uidv}{\mathrm{UID}_{v}\xspace}

\usepackage{xcolor}
\definecolor{ibmyellow}{HTML}{FFB000} 
\definecolor{ibmorange}{HTML}{FE6100}
\definecolor{ibmpurple}{HTML}{785EF0}
\definecolor{ibmblue}{HTML}{648FFF}
\definecolor{ibmred}{HTML}{DC267F}

\definecolor{ibmorangelight}{HTML}{FFF2E6}
\definecolor{ibmpurplelight}{HTML}{F2EEFF}

\definecolor{backT}{HTML}{FF8300}      
\definecolor{backp}{HTML}{785EF0}     
\definecolor{backD}{HTML}{E62325}      
\definecolor{backD}{HTML}{648FFF}     

\newcommand{\U}{\textcolor{ibmorange}{U}\xspace}
\newcommand{\Pp}{\textcolor{ibmpurple}{P}\xspace}
\newcommand{\D}{\textcolor{ibmred}{D}\xspace}
\newcommand{\PD}{$\text{\Pp}+\text{\D}$\xspace}

\newcommand{\DeltaP}{{\Delta_{\text{\Pp}}}}
\newcommand{\DeltaD}{{\Delta_{\text{\D}}}}
\newcommand{\DeltaPD}{{\Delta_{\text{\Pp\!+\D}}}}
\newcommand{\edin}{\Pp}
\newcommand{\ucl}{\D}
\newcommand{\kul}{\U}

\title{Is Information Density Uniform \\ when Utterances are Grounded on Perception and Discourse?}

\author{Matteo Gay$^{\kul, }$\thanks{Research done while visiting the University of Edinburgh.}~~~~~Coleman Haley$^{\edin}$~~~~~Mario Giulianelli$^{\ucl}$~~~~~Edoardo M. Ponti$^{\edin}$\\
$^{\kul}$KU Leuven  \quad $^{\edin}$ University of Edinburgh  \quad $^{\ucl}$ University College London \\
$^{\kul}$\texttt{matteo.gay@student.kuleuven.be} ~~~~~$^{\edin}$\texttt{coleman.haley@ed.ac.uk}\\  $^{\ucl}$\texttt{m.giulianelli@ucl.ac.uk} ~~~~~$^{\edin}$\texttt{eponti@ed.ac.uk}}

\date{}
\begin{document}
\maketitle

\begin{abstract}
The Uniform Information Density (UID) hypothesis posits that speakers are subject to a communicative pressure to distribute information evenly within utterances, minimising surprisal variance. While this hypothesis has been tested empirically, prior studies are limited exclusively to text-only inputs, abstracting away from the perceptual context in which utterances are produced. In this work, we present the first computational study of UID in visually grounded settings. 
We estimate surprisal using multilingual vision-and-language models over image--caption data in 30 languages and visual storytelling data in 13 languages, together spanning 11 families. We find that grounding on perception consistently smooths the distribution of information, increasing both global and local uniformity across typologically diverse languages compared to text-only settings. In visual narratives, grounding in both image and discourse contexts has additional effects, with the strongest surprisal reductions occurring at the onset of discourse units. 
Overall, this study takes a first step towards modelling the temporal dynamics of information flow in ecologically plausible, multimodal language use, and finds that grounded language exhibits greater information uniformity, supporting a context-sensitive formulation of UID.

\end{abstract}

\section{Introduction}
\label{sec:intro}
    
    The Uniform Information Density (UID) hypothesis posits that when language users ``\textit{have a choice between several variants to encode their message, they prefer the variant with more uniform information density (ceteris paribus\textit)}''  \citep{jaeger2010redundancy}. 
    The preference for avoiding sharp fluctuations in information transmission results in a smoothing effect over the ``information contour'' of the signal \citep{tsipidietal2024surprise}, a phenomenon thought to be linked to a reduced cognitive load and more efficient processing \citep{meister-etal-2021-revisiting, clark2023cross}. In practice, this effect is realised through linguistic choices\footnote{As in \citet{jaeger2010redundancy}, ``\textit{the term `choice' does not imply conscious decision making. It is simply used to refer to the existence of several different ways to encode the intended message into a linguistic utterance.}''} that minimise the variance of surprisal across linguistic units (e.g., words) within a specified contextual window (e.g., a sentence or paragraph).\looseness-1

    Information density is commonly measured in terms of \textit{surprisal}, i.e., the bits of information conveyed by a word given its context \citep{shannon,futrell2022inftheory}. Traditional computational and psycholinguistic research on UID has predominantly relied on text-only language models to estimate surprisal distribution, focusing on a system-internal notion of UID that abstracts linguistic sequences from extralinguistic context \citep{genzel2002entropy, ayletTurk2004redundancy, ayletTurk2006syllab, jaegerLevy2006syntax}. Moreover, while some studies have examined the influence of broader linguistic contexts \citep[e.g.,][]{doyleFrank2015twitter, giulianellietal2021information},
    most prior work has treated linguistic units (e.g., sentences) in isolation 
    \citep[e.g.,][]{frank2008speaking, mahowald2013info}.
    In practice, however, linguistic communication often involves the incremental production and interpretation of linguistic material grounded in both perceptual and discourse context. Recent psycholinguistic research finds that this contextual grounding facilitates communication in naturalistic settings \citep[such as face-to-face conversation,][]{drijvers2023facetoface} and in more controlled experimental tasks \citep[such as the Image-Conditioned Maze Task proposed by][]{pushpita-levy-2024-image}.\looseness-1
    
    In this paper, we present---to the best of our knowledge---the first investigation of the UID hypothesis in a grounded setting. Specifically, we estimate linguistic surprisal conditioned on visual stimuli \citep[following][]{haley2024grounded}
    and assess how such grounding influences the distribution of information density. 
    We adopt a typological perspective, including 33 languages from 11 families and 2 macro-areas\footnote{See \cref{tab:APP_langs} in \cref{app:dataset_stats} for the detailed taxonomy.} as defined in \citet{wals}. First, we analyse a dataset of image–caption pairs \citep{haley2024grounded}.
    We then further extend this analysis by jointly examining the roles of perceptual
    and discourse context, drawing on visual storytelling datasets \citep{leong-etal-2022-bloom}. 
    To measure surprisal, we leverage multilingual vision-and-language models. Specifically, we utilise word-level surprisal estimates for image captions derived from PaliGemma \citep{beyer2024paligemmaversatile3bvlm} --- as provided in \textsc{Ground-XM3600} --- and we further extract surprisal values for visual narratives using Gemma 3 \citep{gemmateam2025gemma3technicalreport}.
    This allows us to study, via visual grounding and controlled contextual ablation, the contribution of multimodal information to the dynamics of information flow in texts associated with visual stimuli. 
    
    In particular, we study the following research questions: (1) How is the information density of an utterance shaped by grounding on visual perception? (2) How do discourse and perceptual context interact in modulating information density in visual narratives? (3) How does information density evolve over the time course of visually grounded discourse?
    We find that grounding on images robustly lowers surprisal variance across typologically diverse languages, both locally between adjacent words and globally over entire utterances. In visual narratives, discourse and perceptual context yield additional variance reduction of surprisal, with the strongest smoothing effects concentrated at sentence and paragraph onsets. Uniformity increases over the course of a story, at first abruptly and then slowly, when perception and discourse are considered. Overall, these findings support the view that perceptual and discourse context enables speakers to distribute information more evenly and thereby communicate more efficiently. We release code and data for our experiments at \url{https://github.com/Imatgay/Grounded-UID}.

\section{Background}
\label{sec:background}
    \paragraph{Information, context, structure.} 
    Most computational and psycholinguistic studies of UID have operated in unimodal, text-only regimes. Much of this literature has focused on explaining speakers' syntactic choices---when alternative constructions are available---in terms of a pressure toward UID at the sentence or clause level \citep[e.g.,][]{frank2008speaking, jaeger2010redundancy, gamboa_2024_uid}. 
    At a broader level of granularity, the UID hypothesis has also been employed as a framework for interpreting the dynamics of information flow in extended discourse \citep{genzel2002entropy, keller-2004-entropy,giulianelli-fernandez-2021-analysing,verma-etal-2023-revisiting}. 
    Research on discourse-level information dynamics has frequently intersected with analyses of structural boundaries (e.g., paragraph breaks or conversational turns) and their effect on surprisal distribution, both in monological \citep{genzel-charniak-2003-variation, tsipidietal2024surprise,tsipidi2025harmonicstructureinformationcontours} and dialogical contexts \citep{xu_2018_dialogue, giulianellietal2021information}.

    Nevertheless, existing work on UID and surprisal contours has yet to consider the influence of extralinguistic, multimodal context on information distribution. In addition, the empirical scope of these studies has been typologically narrow, focusing primarily on a limited set of Indo-European languages.
    
    \paragraph{Visual grounding.}
    The role of image-mediated context in shaping linguistic expectations has been explored in two recent studies, both operationalising groundedness as a reduction in word-level surprisal under multimodal conditioning via vision-and-language models. \citet{haley2024grounded} adopt a typological and information-theoretic perspective, treating images as language-agnostic proxies for utterance meaning and defining and measuring \textit{groundedness} 
    across 30 typologically diverse languages. 
    Instead, \citet{pushpita-levy-2024-image} approach the problem from an experimental psycholinguistic angle, treating images as visual stimuli and finding that surprisal reductions derived from four VLMs (trained with distinct alignment objectives) predict human reading times in an \textit{Image-Conditioned Maze Task}. 
    
    These two perspectives are jointly informative for the present study. 
    On the one hand, the surprisal difference operationalised by \citet{haley2024grounded} allows us to treat groundedness dynamics as a signal of \textit{sequence-level contentfulness}. On the other hand, \citet{pushpita-levy-2024-image} situate visual context within a psycholinguistically plausible model of language processing, validating multimodal surprisal as a cognitively interpretable measure of online comprehension effort. Although VLM-based surprisal is still an emerging psychometric proxy requiring further validation, together, these frameworks support our use of VLM-predicted surprisal contours not only as a computational lens on information distribution, but also as a tentative approximation of ecologically grounded language use.
    Unlike the present work, both studies, consistent with their respective aims, abstract away from the \textit{temporal} organisation of surprisal, focusing exclusively on pointwise predictability. 

    \paragraph{Visual World Paradigm (VWP).}
    Psycholinguistic VWP studies consistently demonstrate that comprehenders exploit visual environments to generate anticipatory eye movements in response to unfolding linguistic input (for a summary, see \citealp{huettig_2011_VisualWorldParadigm}). Foundational works show that such visual grounding facilitates syntactic  disambiguation \citep{tanenhaus_1995_integration}, enhances sub-lexical prediction \citep{dahan_2001_timecourse}, and supports the expectation of upcoming referents based on visual scene constraints \citep{altmann_1999_incrementalverbs}. 

    More recently, \citet{ankener_2018_visualsurprisal} show that \textit{multimodal surprisal}---reflecting the integration of visual and linguistic input---predicts cognitive load as measured through pupil dilation and N400 amplitudes, with the surprisal of nouns decreasing under lower visual ambiguity.
    These findings provide experimental evidence that anticipatory and responsive uncertainty during online processing are 
    not determined solely by the linguistic signal, but reflect dynamically updated expectations conditioned on concurrent multimodal context.

    \paragraph{Research gap.}
    Computational psycholinguistics has modelled the temporal dynamics of information flow via surprisal contours, but this work has largely been limited to unimodal, text-only contexts.
    In contrast, experimental studies have examined language processing in more  ecologically plausible, multimodal settings, but they have primarily focused on local facilitation effects at specific lexical targets.
    \textit{As a result, the broader sequential structure of information in multimodal discourse remains empirically unexplored.} 
    This gap is theoretically significant: if multimodal context facilitates language processing, its effects should extend beyond isolated lexical items to shape the global organisation of information across time.

\section{Research Questions and Hypotheses}
  \label{par:res_gap}
    We combine approaches from two strands of psycholinguistic research, computational and experimental, to study information density across multimodal discourse.
    Specifically, using multilingual vision-and-language models, we examine how discourse context and visual grounding jointly shape the temporal distribution of surprisal---first in image--caption pairs, and then across extended narrative segments.
    Our analysis is guided by the following three research questions.
\label{sec:RQs_Hs}
    \paragraph{RQ1: \textit{How is the information density of an utterance shaped by grounding on visual perception?}}
    Image conditioning makes text more easily predictable: this facilitation effect, quantified as reduction in word surprisal, extends beyond relatively straightforwardly grounded words, consistent with the \textit{comprehensive-grounding hypothesis} \citep{pushpita-levy-2024-image, haley2024grounded}. Specifically, visual grounding induces a non-uniform reduction in surprisal across words, generally more pronounced for words traditionally classified as \textit{content} words than for \textit{function} words. Since content words tend to be higher-surprisal in text-only settings, such a decrease should reduce the variance of the surprisal distribution by narrowing its overall spread. As a consequence, we expect that information is more evenly distributed when grounded, yielding a flatter surprisal contour.

    \paragraph{RQ2: \textit{How do discourse and perceptual context interact in modulating the information density in visual narratives?}}
    We examine how contextual grounding---both textual and visual---modulates information density at the paragraph level in visual storytelling. We hypothesise that discourse context, which shapes the common ground to constrain interpretation and topic, and visual context, which anticipates the referential content of an utterance, independently contribute to reducing surprisal variance. Crucially, we posit that their joint availability---offering complementary cues about both past and upcoming content---yields an additional smoothing effect on information density.

    \paragraph{RQ3: \textit{How does information density evolve over narrative time?}}
    Building on recent proposals that information flow in discourse exhibits harmonic and hierarchical structuring \citep{tsipidietal2024surprise, tsipidi2025harmonicstructureinformationcontours}, we examine how the presence of multimodal contextual material modulates surprisal across nested narrative levels (sentences within paragraphs, paragraphs within stories), thereby affecting UID. 
    We hypothesise that: (i) discourse and perceptual context jointly induce a downward drift in surprisal and $\mathrm{UID}$ over narrative time. This would be consistent with the accumulation of contextual constraints that increasingly narrow the hypothesis space for upcoming content, thereby reducing predictive uncertainty and flattening the surprisal contour over time. 
    We also hypothesise that (ii) contextual information---whether in the form of discourse history carrying background knowledge, or visual \textit{semantic priming}---primarily attenuates the local discontinuities in surprisal observed at discourse boundaries, as these transitions introduce \textit{surprising} new referents or thematic shifts \citep{genzel2002entropy, tsipidi2025harmonicstructureinformationcontours} that multimodal conditioning can help to anticipate. These constitute one of the main structure-driven sources of deviation from an optimal instantiation of the UID hypothesis.

\section{Method}
\label{sec:method}

    At the most fundamental level, the method for addressing these questions consists of: (i) computing metrics for information uniformity (UID values) for an utterance 
    under different conditions (e.g., in isolation, conditioned on an image, or conditioned on prior discourse); and (ii) comparing these UID values to assess how the conditioning context influences the distribution of information. We adhere to the classical information-theoretic model, quantifying the information content of a linguistic unit with \textit{surprisal}.\footnote{
        A range of alternative information measures may be suitable for such analyses \citep[][\textit{inter alia}]{rabovsky2018modelling,aurnhammer2019evaluating,giulianelli-etal-2023-information,giulianelli-etal-2024-generalized,giulianelli-etal-2024-incremental,meister-etal-2024-towards,li2024information}; we leave these for future work.
    }

    \paragraph{Surprisal estimation.} 
    We estimate surprisal at the word level with pre-trained autoregressive vision\--and\--language models and language models. Surprisal is defined as the information content of a word $w_t$ given its preceding context. 
    Concretely, for a word $w_t$ in an utterance $\mathbf{u} = (w_1, \dots, w_n)$, we compute the surprisal $s_t$ as $-\log p(w_t \mid w_1, \dots, w_{t-1}, \mathbf{c})$, where $\mathbf{c}$ is some additional context (e.g., perception or discourse). The conditional probability distribution $p(\cdot)$ is provided by the next-word distribution under a given pre-trained model.
    Since this operates on subword units---in our case tokenised via Sentence\-Piece \citep{kudo-richardson-2018-sentencepiece}---we reconstruct word-level surprisal by summing the negative log probabilities of the tokens each word comprises and we adopt the correction for trailing whitespaces introduced by \citet{oh2024leadingwhitespaceslanguagemodels} and \citet{pimentel2024computeprobabilityword}.\footnote{For a broader discussion of these issues and their implications for surprisal estimation, see \citet{giulianelli-etal-2024-proper}.}
    This gives us a vector of surprisals $\mathbf{s}$ corresponding to each word in~$\mathbf{u}$. Word segmentation---when not available---was determined using Stanza \citep{qi-etal-2020-stanza} for all languages except Bengali, for which we use BNLP \citep{sarker2021bnlpnaturallanguageprocessing}.\looseness-1

    \paragraph{UID computation.}
    Once a surprisal vector $\mathbf{s}$ has been estimated, as described above, we compute UID values using two variance-based metrics introduced by \citet{collins2014information}, and later formalised by \citet{meister-etal-2021-revisiting}. The first, given in \cref{eq:UID-sentence}, captures uniformity at the \textit{global} sequence level (e.g., a sentence or a paragraph):
        \begin{equation}  \label{eq:UID-sentence}
            \uidv(\mathbf{s}) \;=\; \frac{1}{n} \sum_{t=1}^{n} \bigl(s_t - \mu\bigr)^2
        \end{equation}
    where $\mu = \frac{1}{n}\sum_{t=1}^{n} s_t$ is the mean surprisal over the $n$ words in the sequence. This metric quantifies the extent to which the surprisal values of individual units (e.g., words) deviate from the overall sequence average. The second metric, defined in \cref{eq:UID-local}, instead focuses on \textit{local} uniformity, quantifying UID as the average change in surprisal between consecutive units within a sequence:
        \begin{equation}  \label{eq:UID-local}
            \uidlv(\mathbf{s}) \;=\; \frac{1}{n-1} \sum_{t=2}^{n} \bigl(s_t - s_{t-1}\bigr)^2
        \end{equation}
    In both cases, lower values indicate more uniform surprisal profiles. 
    We take the local version of UID to be more sensitive to fine-grained grammatical and locality-driven effects in language, whereas the global metric is more useful for analysing broader, discourse-level patterns of information transmission.\looseness-1
    
\subsection{Experimental Setup}
    \label{subsec:exp}

    For an utterance $\mathbf{u}$, we consider these possible contexts: \textbf{Utterance} (\U), where the utterance is evaluated in isolation, without any preceding linguistic or visual context; \textbf{Perception} (\Pp), where the utterance is conditioned on a corresponding image; and \textbf{Discourse} (\D), where the utterance is conditioned on preceding utterances (when available).

    To address our research questions, we compute UID values of textual sequences under four distinct conditions: \U, \Pp, \D, and \PD. The last condition corresponds to the visual storytelling setting (ViSt; \citealp{huang-etal-2016-visual}), in which surprisal is computed for each utterance based on the full available context---i.e., all the preceding textual sequences and the associated images.

    \paragraph{Models and Datasets.}
    We consider two multimodal datasets for our experiments: the \textsc{Ground-XM3600} dataset \citep{haley2024grounded}\footnote{\href{https://osf.io/bdhna/}{https://osf.io/bdhna/}} for image--caption pairs \citep[specifically, the \textsc{Crossmodal-3600} split;][]{thapliyal-etal-2022-crossmodal} and the \textsc{Bloom\-VIST} dataset \citep{leong-etal-2022-bloom}\footnote{\href{https://huggingface.co/datasets/sil-ai/bloom-vist}{https://huggingface.co/datasets/sil-ai/bloom-vist}} for visual storytelling (image-interleaved text). We report the dataset statistics for both in \cref{app:dataset_stats}.
    
    \textsc{\textbf{Ground-XM3600}} provides word-level surprisal values for 213,677 image--caption pairs across 30 typologically diverse languages from 10 families (\cref{tab:XCROSS_datadist}). Each word is annotated with two surprisal values, computed under two different conditions: (i) \Pp with the \texttt{paligemma\--3b\--ft\--coco35\--224} checkpoint of PaliGemma \citep{beyer2024paligemmaversatile3bvlm} as a vision-and-language model (VLM); and (ii) \U starting with the decoder of pre-trained \texttt{paligemma\--3b\--pt\--224} used as a language model (LM), which was then fine-tuned on the captions of COCO35L. This ensures exact overlap between the textual data observed by the VLM and its LM counterpart.
    
    \textsc{\textbf{BloomVIST}} is a split of the Bloom Library \citep{leong-etal-2022-bloom}, containing 11,407 stories structured as interleaved image--paragraph sequences, with a total of 112,080 image--paragraph pairs. An example of a data sample is provided in \cref{fig:surprisal_ex} (Appendix \ref{app:BLOOM_ex}). Of the original 363 languages represented, we include only the 48 with a number of stories sufficiently large for our analyses ($>20$). We truncate stories after the 20th paragraph as the models' surprisal values may become unreliable for exceedingly long sequences. Moreover, to reduce noise and exclude edge cases, we discarded stories containing at least one paragraph with fewer than three words.
    For this dataset, we used Gemma~3 4B pre-trained \citep{gemmateam2025gemma3technicalreport}\footnote{\href{https://huggingface.co/google/gemma-3-4b-pt}{https://huggingface.co/google/gemma-3-4b-pt}} as a unified instantiation of both VLM and LM. This model natively supports both multimodal and unimodal (text-only) input formats, removing the need for additional fine-tuning, and can 
    handle long contexts (up to 128K tokens) with interleaved text-and-image inputs. However, unlike PaliGemma, the composition of Gemma 3's training data mixture remains undisclosed---including the exact languages it covers. Based on empirical heuristics,\footnote{Character-level perplexity over \textsc{BloomVIST} stories was used to contrast our language model with one with randomly initialised embeddings, filtering out languages with moderate and low discrepancies. 
    High perplexity was tolerated in languages with high intrinsic character entropy (e.g., Mandarin Chinese), under the assumption of sufficient representation in the training data.} 
    we ultimately selected the 13 languages (from 7 language families) listed in \cref{tab:BLOOM_datadist}.

    \begin{figure*}[htb!]
  \centering
  \includegraphics[width=\textwidth]{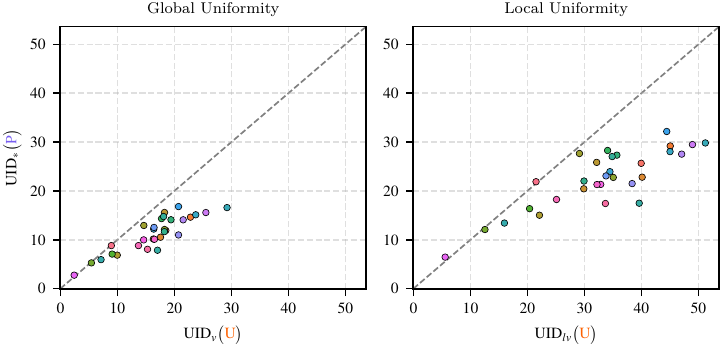}
  \caption{Global and local UID values under the \U\ condition (caption in isolation) and \Pp\ condition (caption with visual context) across the 30 typologically diverse languages, from 10 language families, in \textsc{Ground-XM3600} dataset. Points below the dashed diagonal line indicate increased uniformity (i.e., decreased global or local variability) when the caption's surprisal is conditioned on the image context.}
  \label{fig:xcross_scatterplot}
\end{figure*}

    \paragraph{Metrics.}
    To assess whether visually grounded language exhibits different information contours than text-only language, we examine information uniformity across both \textsc{Ground-XM3600} and \textsc{BloomVIST}. 
    We compute $\uidv$ and $\uidlv$ according to Equations~\eqref{eq:UID-sentence} and~\eqref{eq:UID-local} respectively.
    For \textsc{Ground-XM3600}, each caption is treated as a single utterance. For each language, we aggregate UID metrics across all captions, reporting mean $\uidv$ and $\uidlv$ under the \U and \Pp conditions, and compute their relative difference $\Delta$ to quantify the impact of visual grounding.
    As for \textsc{Bloom\-VIST}, word surprisal estimates are obtained using Gemma~3 under \U, \Pp, \D, and \PD. $\uidv$ is then computed under all four conditions at the paragraph and sentence level.\footnote{Paragraphs consist of text interleaved with images according to the predefined alignment of the dataset. Sentences are extracted with Stanza's sentence tokenizer \citep{qi-etal-2020-stanza}.} To make the comparison fair across conditions where the average density differs, we also report the Coefficient of Variation (CV $= \frac{\sigma(\mathbf{s})}{\mu(\mathbf{s})}$), which is a unitless (scale-free) measure of how uneven the densities are.
    
    To investigate how discourse and perceptual context modulate the organisation of information throughout a narrative, we adopted two complementary measures: (i) linear mixed-regression modelling of information-theoretic metrics against relative position within discourse; and (ii) density estimation of surprisal reductions over normalised narrative positions induced by each kind of context.
    First, adapting the methodology of \citet{giulianelli-fernandez-2021-analysing}, we fitted linear mixed-effect models for each language and level of granularity (sentence or paragraph), predicting surprisal and $\uidv$ as a function of the relative position of a unit within the immediately higher structural level (in our case, sentence~<~paragraph~<~story), under different context conditions (\U, \Pp, \D, \PD). The resulting slopes coarsely quantify the drift of information flow under different contexts.
    Second, we examined the distribution of positive surprisal reductions. For each word, we computed three reduction scores ($\DeltaP$, $\DeltaD$, $\DeltaPD$), reflecting the contextual facilitation due to local perceptual, discourse, and joint multimodal context, respectively. 

\section{Results}
    \subsection{How does Visual Grounding Shape Utterance Information Density?}
    First, to quantify the impact of visual grounding on textual information distribution, we compared sentence-level UID values---both global ($\uidv$) and local ($\uidlv$)---across 30 languages in the \textsc{Ground-XM3600} dataset, as described in \cref{subsec:exp}. For every caption, we computed UID from word-level surprisal under \U\ and \Pp\ conditions. We then performed paired Wilcoxon signed-rank tests \citep{wilcoxon1945} per language and UID metric to assess the statistical significance of UID changes due to grounding, while computing standardised effect sizes (paired Cohen’s $d_z$). 
    Results are shown in \cref{fig:xcross_scatterplot} (for complete data, see \cref{tab:XCROSS_deltas} in Appendix \ref{app:XCROSS_ rq1}). \looseness-1

    \begin{figure}[t]
  \centering
  \includegraphics[width=\columnwidth]{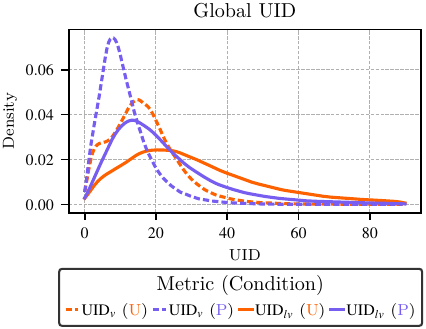}
  \caption{Kernel density estimation of global UID values for captions in \textsc{Ground-XM3600}, conditioned either on their respective image (\Pp) or without any contextual image (\U). Curves are truncated at the 99th percentile to enhance visual clarity.}
  \label{fig:XCROSS_uid_distribution}
\end{figure}
     \begin{figure*}[t]
    \centering
    \includegraphics[width=\textwidth]{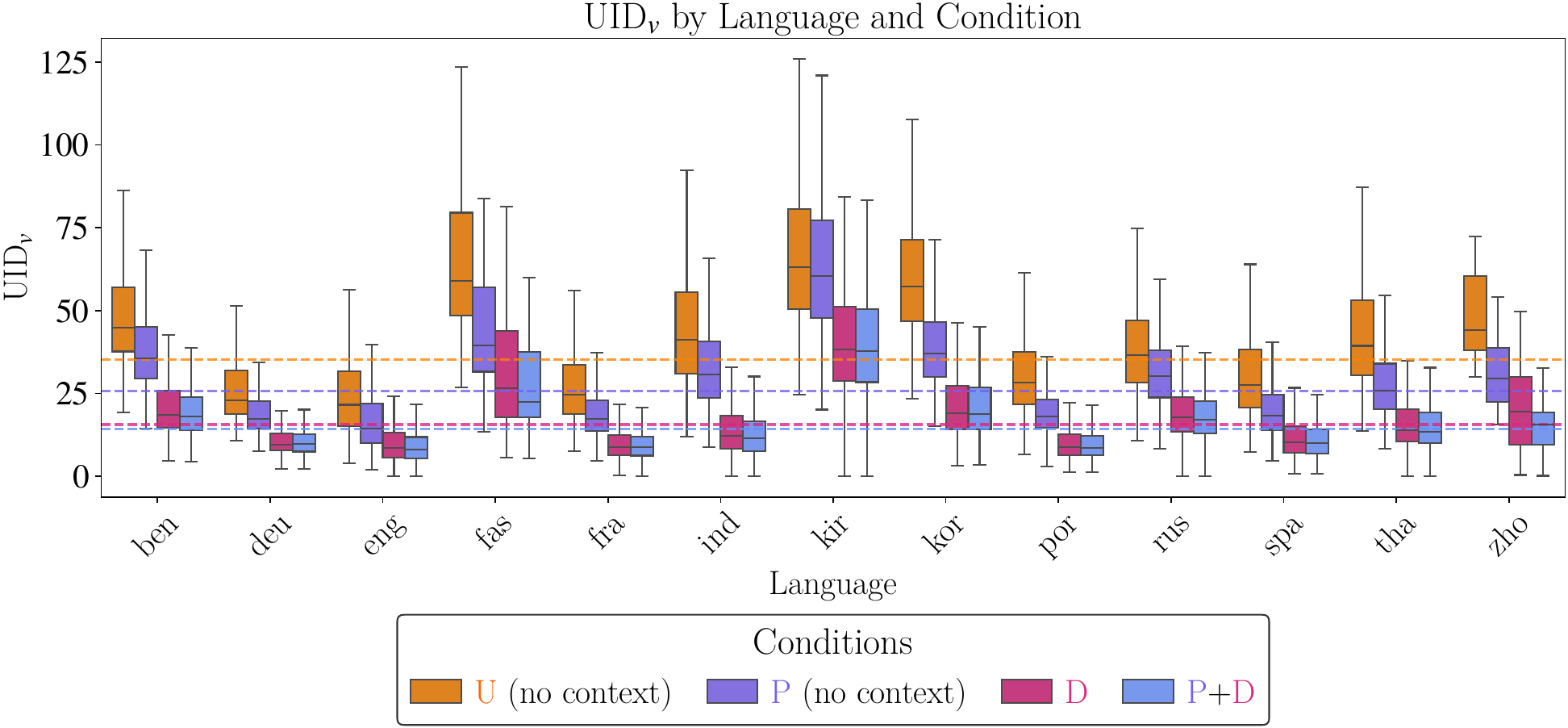}
    \caption{Each box represents the distribution of paragraph-level $\uidv$ values across languages under four conditions:
    no context (\U);
    paired image as context (\Pp);
    all preceding paragraphs as context (\D);
    all preceding paragraphs and interleaved images as context (\PD). Dashed horizontal lines indicate the mean $\uidv$ per condition, averaged across all languages.}
    \label{fig:BLOOM_boxplot}
\end{figure*}

    The hypothesis formulated in \cref{sec:RQs_Hs} is confirmed in virtually all languages, for both local and global UID metrics: grounding in visual perception reliably reduces UID values, indicating a more uniform information distribution across utterances. 
    Exceptions are limited to Telugu, where both $\uidv$ and $\uidlv$ significantly increase when grounded, and Arabic, where only $\uidlv$ does.
    We suspect that these anomalous cases may reflect limitations in the model's vision--language alignment or be influenced by script-specific and morphological properties that affect token segmentation \citep{park-etal-2021-morphology, petrov2023language, oh-schuler-2025-impact}.
    It should be further noted that UID decreases in all other languages by different degrees, possibly for similar reasons.
    The shift induced by grounding is further illustrated in \cref{fig:XCROSS_uid_distribution}, which shows the estimated density of UID values, averaged across all sentences and languages. 
    For both metrics, the density curves under visual grounding (\Pp) are consistently shifted leftwards compared to their text-only (\U) counterparts, with a more pronounced concentration of low UID values.

    \subsection{How does the Interaction of Discourse and Perceptual Context Modulate Information Density?}
    Secondly, we extend our analysis to the \textsc{Bloom\-VIST} dataset to include discourse context (\D). 
    Specifically, for every image--paragraph pair within a story, we estimated UID metrics under the four conditions listed in \cref{subsec:exp}, as shown in
    Figure~\ref{fig:BLOOM_boxplot}. A clear bifurcation emerges between conditions without prior discourse (\U, \Pp) and those incorporating it (\D, \PD), with the latter group exhibiting consistently lower median UID values. Within each group, visually grounded conditions (\Pp\ and \PD) yield lower UID values than their text-only counterparts (\U\ and \D), indicating a robust contribution of perceptual context across discourse context regimes. Notably, the joint condition \PD\ on average yields the lowest UID values, and its distribution closely tracks that of \D. This indicates that discourse context accounts for a relatively larger reduction in UID, while perceptual context contributes a small but consistent additional smoothing when both contextual modalities are available.
    To statistically test for the hypothesised monotonic ordering of UID values $(\text{\U} > \text{\Pp} > \text{\D} > \text{\Pp}+\text{\D})$, we applied the Page test for ordered alternatives \citep{Page1963rankedtest} independently for each language. The test yielded highly significant results across all languages ($p < 0.001$), providing robust support for the hypothesised hierarchy of contextual effects on UID.\looseness-1

    \begin{figure}[t!]
  \centering
  \includegraphics[width=\columnwidth]{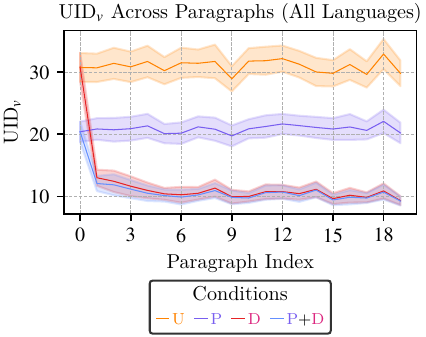}
  \caption{Paragraph-level UID values averaged across \textsc{BloomVIST} stories (truncated at 20 paragraphs) under the utterance only (\U), perceptual context (\Pp), discourse context (\D), and multimodal context (\PD) conditions.}\label{fig:BLOOM_20_paragraphs}
\end{figure}

    \subsection{How do Visually Grounded Information Contours Evolve over Narrative Time?}
    
    Figure~\ref{fig:BLOOM_20_paragraphs} plots mean $\uidv$ for each paragraph index in visual stories with a maximum of 20 paragraphs, under the four conditions we study. UID trajectories under \U\ and \Pp\ are relatively flat, whereas discourse-informed conditions (\D, \PD) show a marked drop over the early paragraphs and then slowly flatten out. The curves for \D and \PD get increasingly closer as more discourse context accumulates. This suggests that the textual history eventually encodes as much information about a paragraph as the image paired with it.
    
    Fitting linear mixed-effects models of UID and mean surprisal against relative (normalised within-unit) position at the paragraph and sentence level yields slope values, which help us determine the evolution of information flow under different conditions. These slopes are extracted from the interaction between position and context condition, with \U\ serving as the baseline. In the \U\ (no context) and \Pp (image context) conditions, surprisal and $\uidv$ tend to drift negatively over narrative time at the sentence level, but show weaker or inconsistent trends at the paragraph level, with most paragraph-level slopes being statistically non-significant (see \cref{tab:BLOOM_mixed_surprisal,tab:BLOOM_mixed_uidv} in Appendix~\ref{app:BLOOM_ols}). This indicates that without conditioning information estimates on global discourse context, no meaningful discourse-level patterns can be observed. 
    
    In contrast, the introduction of discourse (\D) and multimodal (\PD) context conditions induces consistent and significantly negative slopes at both sentence and paragraph levels across languages. This suggests that mean surprisal and UID scores of discourse units decrease as constraints accumulate over the course of the narrative. 
    Notably, the negative slope becomes systematically less steep when images are added to discourse (\PD vs \D), suggesting that additional contextual information leads to higher global stability throughout the paragraph or story.\looseness-1

    This global flattening effect raises the question of \textit{where} within discourse units contextual information exerts its strongest impact. If surprisal is primarily concentrated at unit onsets, as \citet{tsipidi2025harmonicstructureinformationcontours} suggest, we might expect the observed variance reduction to be driven, in large part, by the local smoothing of those unit-initial surprisal peaks.
    Across all plotted languages and levels (sentence and paragraph), surprisal reductions attributable to perceptual ($\DeltaP$), discourse ($\DeltaD$), and combined context ($\DeltaPD$) exhibit a consistent front-loaded distribution (see \cref{app:density_plot}), peaking sharply at the onset of the unit and decaying rapidly thereafter.
    This pattern indicates that contextual information---whether visual, textual, or multimodal---exerts the strongest disambiguating and constraining effect at the beginning of sentences and paragraphs. The effect is especially pronounced for discourse context ($\DeltaD$), which shows the highest density near position zero in nearly all cases. Multimodal context ($\DeltaPD$) behaves similarly but tends to yield smoother reductions with longer tails, suggesting more sustained facilitation throughout the unit.

\section{Further Discussion}

\subsection{A Scale-free Metric of Uniformity}
\label{subsec:cv_analysis}
    
    To account for possible discrepancies in mean surprisal across conditions, which may affect variance, we also computed the Coefficient of Variation as a scale-free measure of dispersion, as described in \cref{subsec:exp}. The results of this analysis are reported in \cref{tab:xcross_coeffvar,tab:bloom_coeffvar}, \cref{app:cv}. 
    In \textsc{Ground-XM3600}, visual grounding significantly increases CV across all languages ($p < .001$), in contrast with the observed reduction in UID values. 
    This is most likely due to $\mu$ decreasing proportionally more than $\sigma$ in short captions, where images collapse non-onset surprisals. For instance, in \textit{``A polar bear is swimming''}, \U surprisals $[10.34, 7.87, 0.08, 0.98, 2.95]$ ($\mu = 4.44, \sigma = 4.46, CV = 1.01$) shift under \Pp to $[10.45, 0.49, 0.01, 1.43, 0.39]$ ($\mu = 2.55, \sigma = 4.44$), yielding $CV = 1.74$. Such behaviour suggests the metric may be less reliable than raw variance when predicting processing effort in short-form referential language conditioned on images. For this reason, we think the interplay between alternative dispersion metrics and standard UID measures warrants further systematic investigation in the future.
    
    Conversely, \textsc{BloomVIST} exhibits more coherent dynamics across standard UID and scale-free metrics, with grounding typically preserving or reducing CV. Unlike isolated captions, narrative paragraphs involve complex dynamic discourse structures where visual context disambiguates the semantic content in a more distributed fashion, preventing the collapse of $\mu$.

\subsection{Linguistic Analysis}
    \label{subsec:linguistic_drivers}

    Finally, we propose a simple linguistic analysis of instances of UID reduction failure under perceptual grounding---i.e., cases in which conditioning on perceptual context does not lead to a reduction in UID. By leveraging Part-of-Speech (POS) annotations in \textsc{Ground-XM3600}, we decompose $\uidv$ into word-level variance contributions across POS categories.
    This allows us to quantify how the contribution of each POS category changes when utterances are grounded in visual perception (\Pp) relative to text-only conditions (\U). 
    We focus on cases in which perceptual grounding increases sentence-level variance, despite the overall tendency for grounding to reduce UID.
    
    The cross-linguistic heatmap (\cref{fig:XCROSS_heatmap_distribution} in Appendix~\ref{app:pos_analysis}) visualises the average change in these variance contributions for sentences where $\uidv(\text{\Pp}) > \uidv(\text{\U})$. Across typologically diverse languages, Proper Nouns (PROPN), Numerals (NUM), and Adjectives (ADJ) emerge as the primary drivers of non-uniformity (that is, words whose surprisal under \Pp deviated strongly from the sentence-level mean). As further shown in \cref{app:pos_analysis}, higher variance contributions under \Pp relative to \U typically coincide with an increase in absolute surprisal for these POS categories.

    These findings warrant further investigation and raise questions regarding the role of specific POS categories in UID and the robustness of VLMs as grounded surprisal estimators.
    For instance, in vision–language models, adjectives are often the main remaining source of uncertainty. In line with \citet{buettner2024investigating}'s notion of ``attribute insensitivity'' at the representation level, VLMs may recognise referents in the image better than their attributes. 
    
\section{Conclusions and Future Work}

    We evaluated how grounding utterances on visual perception and discourse affects information contours. Leveraging multilingual vision-and-language models and established information-theoretic metrics, we studied datasets of captioned images (30 languages spanning 10 families) and visual storytelling (13 languages, 7 families).

    Our findings indicate that when utterances are conditioned on perceptual context, information is distributed more evenly across the textual sequence than when utterances are considered in isolation, both in image captions and in visually grounded stories. 
    In visual storytelling, discourse context alone induces a substantial smoothing of the surprisal contour, yet the incorporation of visual context to discourse yields a consistent, albeit small, additional reduction in surprisal variance. In light of recent accounts of information dynamics in extended discourse---such as the Structured Context hypothesis \citep{tsipidietal2024surprise} and the Harmonic Surprisal hypothesis \citep{tsipidi2025harmonicstructureinformationcontours}---we further suggest that surprisal reduction from grounding exhibits systematic structure across scales: contextual information exerts its strongest effect at unit onsets, both at the sentence and paragraph level. This strongly front-loaded profile indicates that discourse and perceptual grounding primarily enhance UID acting at structural boundaries, which we identify as key loci for contextual integration in the regulation of information flow.
    
    Moreover, building on \citet{haley2024grounded}'s interpretation of images as ``\textit{language-agnostic representations of meaning}'',
    we extended their line of inquiry beyond static image--caption pairs---where captions are typically tightly and referentially linked to their associated images---to the looser, more anticipatory image–paragraph pairings found in visual storytelling. In this latter setting, the relationship between image and text is not strictly referential but often reflects broader narrative structure and event dynamics. Within this framework, we find suggestive evidence that images can convey higher-level discourse-semantic content, thereby modulating the global information distribution of subsequent utterances. This points to a more abstract form of visual grounding, in which perceptual context contributes not only to local lexical predictability but also to the shaping of information flow at the level of discourse structure.
    
    A natural extension of the present work is to move beyond static images toward dynamic perceptual streams, such as video, where the temporal structure of the input can more tightly constrain linguistic predictions. Unlike static images, dynamic stimuli unfold over time and encode event-level structure (e.g., causality, agency, temporal progression), potentially offering a richer basis for modelling how perceptual context informs discourse organisation. Additionally, these findings suggest the need for formal decomposition into redundant and synergistic contributions (e.g., via Partial Information Decomposition; \citealp{luppi_2024_informationdecomposition}), enabling a more precise characterisation of how modalities combine. Finally, inspired by \citet{pushpita-levy-2024-image}, it would be valuable to complement our computational modelling analysis with a tailored behavioural experiment designed to probe the cognitive correlates of the effects we observe.

\section*{Limitations}
This study is subject to three main limitations.
First, while we adopt a typologically informed approach, the intersection between available datasets and model capabilities effectively constrains our choice of languages. In particular, although \textsc{Bloom\-VIST} provides coverage of several under-resourced languages, most of these fall outside the training distribution of Gemma 3, resulting in a final language set heavily biased towards Indo-European languages and limited to two WALS macro-areas (Eurasia and Africa, per \citealp{wals}). Second, our analysis is computational, without psycholinguistic validation: while model-based surprisal estimates are supported by recent studies \citep{pushpita-levy-2024-image}, no experimental data yet links these predictions to actual human processing patterns in multimodal discourse; this underscores the need for complementary human studies to validate our findings. Third, the current availability of open-weight multilingual VLMs with long-context processing capabilities remains very limited, as does the availability of multilingual visual storytelling datasets. This restricts the ability to test the robustness of our findings across different models and data sources.

\section*{Acknowledgements}
Matteo Gay acknowledges the School of Advanced Studies IUSS Pavia for funding his research visit to the University of Edinburgh, where he conducted the research presented in this paper.

\bibliography{anthology,custom}

\clearpage

\clearpage

\appendix
\section*{Appendix}

\section{Dataset Statistics}
    \label{app:dataset_stats}
        Dataset statistics for \textsc{Ground-XM3600} (\cref{tab:XCROSS_datadist}) and \textsc{BloomVIST} (\cref{tab:BLOOM_datadist}). \Cref{tab:APP_langs} provides the full list of the 33 languages categorised by WALS family and macro-area.
        \begin{table}[h]
\centering
\begin{tabular}{>{\scshape}l|rrrr}
\toprule
{\normalfont \textbf{Lang}} & \textbf{Caps} & \textbf{Words} & \textbf{Mean} & \textbf{Std} \\
\midrule
arb & 6928 & 50225 & 7.2 & 3.3 \\
ces & 5867 & 38299 & 6.5 & 5.0 \\
dan & 6696 & 58845 & 8.8 & 4.4 \\
deu & 8615 & 93753 & 10.9 & 3.8 \\
ell & 6040 & 48281 & 8.0 & 4.0 \\
eng & 7179 & 67956 & 9.5 & 3.9 \\
fas & 7142 & 85867 & 12.0 & 6.3 \\
fin & 6726 & 51444 & 7.6 & 4.2 \\
fra & 8520 & 107899 & 12.7 & 4.6 \\
heb & 6572 & 74965 & 11.4 & 8.2 \\
hin & 8501 & 114194 & 13.4 & 5.2 \\
hrv & 6555 & 59564 & 9.1 & 4.9 \\
hun & 6711 & 60706 & 9.0 & 5.5 \\
ind & 7115 & 95210 & 13.4 & 5.5 \\
ita & 8448 & 100852 & 11.9 & 4.0 \\
jpn & 7067 & 109036 & 15.4 & 8.7 \\
kor & 7522 & 68462 & 9.1 & 4.2 \\
nld & 7494 & 59313 & 7.9 & 3.7 \\
nor & 7014 & 62969 & 9.0 & 4.2 \\
pol & 6999 & 64241 & 9.2 & 4.7 \\
por & 7023 & 71809 & 10.2 & 5.8 \\
ron & 7048 & 112008 & 15.9 & 10.3 \\
rus & 7136 & 75043 & 10.5 & 5.2 \\
spa & 8597 & 77948 & 9.1 & 3.1 \\
swe & 6514 & 54298 & 8.3 & 3.9 \\
tel & 7195 & 53807 & 7.5 & 2.0 \\
tur & 6849 & 63026 & 9.2 & 6.5 \\
ukr & 7061 & 70935 & 10.0 & 5.9 \\
vie & 6699 & 100610 & 15.0 & 6.9 \\
zho & 5834 & 84401 & 14.5 & 10.3 \\
\bottomrule
\end{tabular}
\caption{Summary statistics of caption lengths across languages in the \textsc{Ground-XM3600} dataset. \textit{Caps}: number of captions; \textit{Words}: total word count; \textit{Mean} and \textit{Std}: average and standard deviation of caption lengths (in words).}
\label{tab:XCROSS_datadist}
\end{table}
        \begin{table}[h]
\centering
\setlength{\tabcolsep}{4pt} 
\begin{tabular}{>{\scshape}l@{\hskip 6pt}r@{\hskip 4pt}r@{\hskip 4pt}r@{\hskip 6pt}r@{\hskip 4pt}r}
\toprule
{\normalfont \textbf{Lang}} & \textbf{Stories} & \textbf{Pars} & \textbf{Words} & \textbf{W/P} & \textbf{W/S} \\
\midrule
ben & 217 & 1919 & 58198 & 30.33 & 268.19 \\
deu & 21 & 220 & 10827 & 49.21 & 515.57 \\
eng & 1997 & 19602 & 766901 & 39.12 & 384.03 \\
fas & 125 & 599 & 8128 & 13.57 & 65.02 \\
fra & 294 & 3444 & 136148 & 39.53 & 463.09 \\
ind & 225 & 1774 & 30659 & 17.28 & 136.26 \\
kir & 265 & 2679 & 109642 & 40.93 & 413.74 \\
kor & 129 & 2114 & 44233 & 20.92 & 342.89 \\
por & 143 & 2270 & 68755 & 30.29 & 480.80 \\
rus & 262 & 2818 & 142974 & 50.74 & 545.70 \\
spa & 426 & 4275 & 134219 & 31.40 & 315.07 \\
tha & 259 & 2762 & 89090 & 32.26 & 343.98 \\
zho & 34 & 221 & 2796 & 12.65 & 82.24 \\
\bottomrule
\end{tabular}
\caption{\textsc{BloomVist} data: number of stories, paragraphs (Pars), words, mean words per paragraph (W/P), and words per story (W/S).}
\label{tab:BLOOM_datadist}
\end{table} 
        \renewcommand{\arraystretch}{1.15} 

\begin{table}[b]
\centering

\begin{tabularx}{0.95\columnwidth}{l l>{\raggedright\arraybackslash}X}
\hline
\textbf{Macro-area} & \textbf{WALS Family} & \textbf{ISO} \\
\hline
\makecell[l]{\textbf{Africa}\\\textbf{(+ Eurasia)}} & Afro-Asiatic & ARB \\
\hline
\textbf{Eurasia} & Afro-Asiatic & HEB \\
\hline
 & Indo-European & BEN, CES, DAN, DEU, ELL, ENG, FAS, FRA, HIN, HRV, ITA, NLD, NOR, POL, POR, RON, RUS, SPA, SWE, UKR \\
\hline
 & Uralic & FIN, HUN \\
\hline
 & Turkic & TUR, KIR \\
\hline
 & Sino-Tibetan & ZHO \\
\hline  
 & Austro-Asiatic & VIE \\
\hline
 & Austronesian & IND \\
\hline
 & Tai-Kadai & THA \\
\hline
 & Dravidian & TEL \\
\hline
 & Japanese & JPN \\
\hline
 & Korean & KOR \\
 \hline
\end{tabularx}
\caption{Classification of the 33 analysed languages by macro-area and family, as defined in WALS (\citealp{wals}).}
\label{tab:APP_langs}
\end{table}
    
    \clearpage

\section{\textsc{BLOOMVist} Example}
    \label{app:BLOOM_ex}
    \begin{figure*}[]
    \centering
    \includegraphics[width=\textwidth]{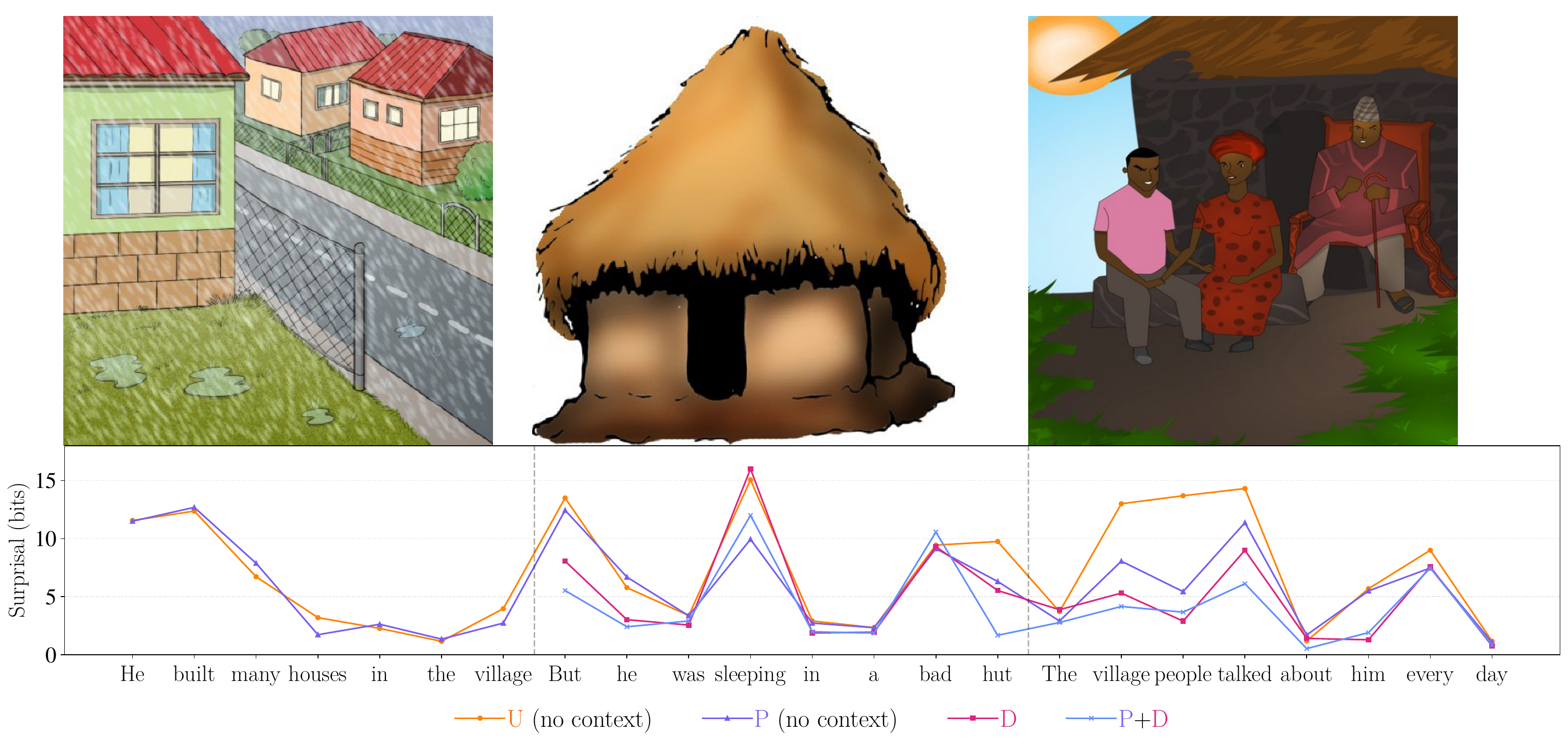}
    \caption{Surprisal contours for the first three paragraphs of a \textsc{BloomVIST} story (ID: dd0b0ef6-3889-47c7-b5ca-93e18e76aba8) across all four experimental conditions. Dashed vertical lines denote paragraph boundaries.}
    \label{fig:surprisal_ex}
\end{figure*}

    \cref{fig:surprisal_ex} provides an example illustration of the surprisal dynamics in a visual narrative from the \textsc{BloomVist} dataset. As a qualitative illustration, observe that concrete referential terms such as \textit{hut} exhibit substantial surprisal reductions when conditioned on visual context. Conversely, abstract or non-referential descriptors like \textit{sleeping} (where the action is not explicitly depicted) and \textit{bad} (a qualification not strictly entailed \textit{a priori} by the image) retain relatively high surprisal.

\section{Grounding Effect in \textsc{Ground-XM3600}}
    \label{app:XCROSS_ rq1}
        Table~\ref{tab:XCROSS_deltas} reports the full language-level results for RQ1, detailing $\uidv$ and $\uidlv$ values in the \textsc{Ground-XM3600} dataset under text-only and visually grounded conditions.
        \begin{table*}[h]
\centering
\begin{tabular}{c|>{\columncolor{ibmorangelight}} r:>{\columncolor{ibmpurplelight}}r|>{\columncolor{ibmorangelight}}r:>{\columncolor{ibmpurplelight}}r|>{\columncolor{white}}r:>{\columncolor{white}}r}
\hline
Lang & UID$_v$ (\U{}) & UID$_v$ (\Pp{}) & UID$_{lv}$ (\U{}) & UID$_{lv}$ (\Pp{}) & $\Delta_v$ (\%) ($d$) & $\Delta_{lv}$ (\%) ($d$) \\
\hline
arb & 8.92 & 8.79 & 21.50 & 21.87 & -1.44~* (-0.02) & \textbf{1.75}~*** (0.02) \\
ces & 18.48 & 11.89 & 39.96 & 25.64 & -35.67~*** (-0.72) & -35.83~*** (-0.54) \\
dan & 22.81 & 14.61 & 45.06 & 29.19 & -35.93~*** (-0.93) & -35.22~*** (-0.64) \\
deu & 17.55 & 10.51 & 40.13 & 22.80 & -40.09~*** (-0.95) & -43.18~*** (-0.87) \\
ell & 16.30 & 10.13 & 29.88 & 20.42 & -37.89~*** (-0.79) & -31.66~*** (-0.53) \\
eng & 18.25 & 15.57 & 32.14 & 25.83 & -14.70~*** (-0.35) & -19.63~*** (-0.31) \\
fas & 9.96 & 6.83 & 22.10 & 15.01 & -31.45~*** (-0.59) & -32.11~*** (-0.49) \\
fin & 14.62 & 12.91 & 29.12 & 27.65 & -11.64~*** (-0.21) & -5.05~*** (-0.07) \\
fra & 18.25 & 12.08 & 35.06 & 22.75 & -33.82~*** (-0.92) & -35.11~*** (-0.72) \\
heb & 5.42 & 5.25 & 12.53 & 12.08 & -3.27~n.s. (-0.04) & -3.59~n.s. (-0.04) \\
hin & 9.06 & 7.05 & 20.37 & 16.35 & -22.21~*** (-0.48) & -19.72~*** (-0.43) \\
hrv & 17.73 & 14.31 & 34.06 & 28.27 & -19.26~*** (-0.42) & -17.02~*** (-0.25) \\
hun & 19.39 & 14.08 & 35.69 & 27.31 & -27.40~*** (-0.61) & -23.47~*** (-0.38) \\
ind & 18.23 & 11.65 & 29.92 & 22.01 & -36.10~*** (-1.05) & -26.44~*** (-0.53) \\
ita & 18.09 & 14.74 & 34.86 & 27.02 & -18.53~*** (-0.54) & -22.49~*** (-0.48) \\
jpn & 17.02 & 7.84 & 39.61 & 17.47 & -53.93~*** (-1.37) & -55.88~*** (-1.17) \\
kor & 7.11 & 5.91 & 15.95 & 13.40 & -16.97~*** (-0.23) & -16.00~*** (-0.21) \\
nld & 29.23 & 16.57 & 51.22 & 29.81 & -43.31~*** (-1.28) & -41.80~*** (-0.84) \\
nor & 23.73 & 15.10 & 45.02 & 28.03 & -36.38~*** (-1.00) & -37.74~*** (-0.70) \\
pol & 16.36 & 12.17 & 34.47 & 23.93 & -25.65~*** (-0.58) & -30.59~*** (-0.52) \\
por & 20.71 & 16.78 & 44.44 & 32.16 & -18.97~*** (-0.49) & -27.64~*** (-0.54) \\
ron & 16.41 & 12.51 & 33.78 & 23.06 & -23.79~*** (-0.61) & -31.72~*** (-0.65) \\
rus & 20.71 & 10.97 & 38.37 & 21.49 & -47.06~*** (-1.36) & -44.00~*** (-0.90) \\
spa & 21.53 & 14.07 & 47.06 & 27.52 & -34.64~*** (-0.92) & -41.51~*** (-0.83) \\
swe & 25.51 & 15.56 & 48.96 & 29.48 & -38.99~*** (-1.00) & -39.79~*** (-0.70) \\
tel & 2.40 & 2.73 & 5.55 & 6.43 & \textbf{13.86}~*** (0.17) & \textbf{15.89}~*** (0.17) \\
tur & 14.59 & 9.96 & 32.85 & 21.32 & -31.73~*** (-0.59) & -35.08~*** (-0.52) \\
ukr & 16.48 & 10.08 & 32.20 & 21.29 & -38.84~*** (-0.69) & -33.87~*** (-0.34) \\
vie & 13.68 & 8.79 & 25.09 & 18.23 & -35.73~*** (-0.90) & -27.35~*** (-0.54) \\
zho & 15.27 & 8.03 & 33.68 & 17.41 & -47.43~*** (-0.98) & -48.29~*** (-0.80) \\
\hline
\end{tabular}
\caption{UID values of \textbf{Ground-XM3600} across languages under \U{} and \Pp{} conditions. The $\Delta$ columns report the relative change in UID from \U{} to \Pp{}, computed as $\frac{(\text{\Pp{}} - \text{\U{}})}{\text{\U{}}}$ and Cohen's $d$ effect size in parentheses. Bold values mark languages where UID is higher under \Pp{}, indicating lower uniformity in textual information when grounded in perception. Statistical significance is based on paired Wilcoxon signed-rank tests across sentences, with Benjamini--Hochberg FDR correction across languages. Significance thresholds: $^\ast$ $q<0.05$, $^{\ast\ast}$ $q<0.01$, $^{\ast\ast\ast}$ $q<0.001$, n.s.~(not significant).}
\label{tab:XCROSS_deltas}
\end{table*}

\section{Coefficient of Variation}
    \label{app:cv}

        In \cref{tab:xcross_coeffvar} (\textsc{Ground-XM3600}) and \cref{tab:bloom_coeffvar} (\textsc{BloomVIST}), we report the results for the scale-free dispersion metric (Coefficient of Variation, CV) discussed in \cref{subsec:cv_analysis}.
        \begin{table}[t]
\centering
\setlength{\tabcolsep}{3.5pt}
\begin{tabular}{l|rr>{\columncolor{gray!10}}r|rr>{\columncolor{gray!10}}r}
\toprule
\textbf{Lang} & \multicolumn{3}{c}{\textbf{Global CV}} & \multicolumn{3}{c}{\textbf{Local CV}} \\
 & \U{} & \Pp{} & $\Delta\%$ & \U{} & \Pp{} & $\Delta\%$ \\
\midrule
arb & 0.52 & 0.75 & $\textbf{43.45}^{***}$ & 0.73 & 1.05 & $\textbf{45.16}^{***}$ \\
ces & 0.72 & 0.89 & $\textbf{23.68}^{***}$ & 0.91 & 1.13 & $\textbf{24.53}^{***}$ \\
dan & 0.80 & 0.98 & $\textbf{22.49}^{***}$ & 1.00 & 1.23 & $\textbf{23.38}^{***}$ \\
deu & 0.71 & 0.86 & $\textbf{21.73}^{***}$ & 1.00 & 1.19 & $\textbf{18.50}^{***}$ \\
ell & 0.71 & 0.81 & $\textbf{12.91}^{***}$ & 0.87 & 1.03 & $\textbf{18.64}^{***}$ \\
eng & 0.82 & 1.10 & $\textbf{34.71}^{***}$ & 0.98 & 1.28 & $\textbf{30.24}^{***}$ \\
fas & 0.64 & 0.80 & $\textbf{24.23}^{***}$ & 0.89 & 1.10 & $\textbf{23.50}^{***}$ \\
fin & 0.58 & 0.77 & $\textbf{33.67}^{***}$ & 0.72 & 1.01 & $\textbf{39.16}^{***}$ \\
fra & 0.80 & 1.03 & $\textbf{27.79}^{***}$ & 1.04 & 1.31 & $\textbf{26.60}^{***}$ \\
heb & 0.52 & 0.77 & $\textbf{48.65}^{***}$ & 0.72 & 1.07 & $\textbf{48.67}^{***}$ \\
hin & 0.68 & 0.79 & $\textbf{15.98}^{***}$ & 0.97 & 1.13 & $\textbf{15.87}^{***}$ \\
hrv & 0.68 & 0.89 & $\textbf{30.28}^{***}$ & 0.84 & 1.12 & $\textbf{32.81}^{***}$ \\
hun & 0.65 & 0.83 & $\textbf{27.12}^{***}$ & 0.79 & 1.03 & $\textbf{30.62}^{***}$ \\
ind & 0.74 & 0.89 & $\textbf{20.71}^{***}$ & 0.89 & 1.15 & $\textbf{29.89}^{***}$ \\
ita & 0.72 & 0.99 & $\textbf{36.03}^{***}$ & 0.93 & 1.24 & $\textbf{33.08}^{***}$ \\
jpn & 0.90 & 1.00 & $\textbf{11.46}^{***}$ & 1.29 & 1.41 & $\textbf{9.50}^{***}$ \\
kor & 0.63 & 0.82 & $\textbf{30.89}^{***}$ & 0.87 & 1.14 & $\textbf{30.61}^{***}$ \\
nld & 0.91 & 1.06 & $\textbf{16.91}^{***}$ & 1.08 & 1.27 & $\textbf{17.86}^{***}$ \\
nor & 0.85 & 1.01 & $\textbf{18.13}^{***}$ & 1.06 & 1.24 & $\textbf{17.38}^{***}$ \\
pol & 0.64 & 0.91 & $\textbf{42.78}^{***}$ & 0.84 & 1.16 & $\textbf{37.43}^{***}$ \\
por & 0.78 & 1.04 & $\textbf{32.35}^{***}$ & 1.03 & 1.29 & $\textbf{25.22}^{***}$ \\
ron & 0.73 & 0.91 & $\textbf{24.23}^{***}$ & 0.98 & 1.15 & $\textbf{17.66}^{***}$ \\
rus & 0.71 & 0.92 & $\textbf{30.22}^{***}$ & 0.88 & 1.19 & $\textbf{34.12}^{***}$ \\
spa & 0.84 & 1.03 & $\textbf{22.05}^{***}$ & 1.12 & 1.30 & $\textbf{16.39}^{***}$ \\
swe & 0.84 & 1.02 & $\textbf{22.13}^{***}$ & 1.03 & 1.25 & $\textbf{21.94}^{***}$ \\
tel & 0.45 & 0.65 & $\textbf{42.06}^{***}$ & 0.64 & 0.90 & $\textbf{42.25}^{***}$ \\
tur & 0.61 & 0.76 & $\textbf{26.18}^{***}$ & 0.82 & 1.01 & $\textbf{23.58}^{***}$ \\
ukr & 0.66 & 0.79 & $\textbf{18.69}^{***}$ & 0.84 & 1.05 & $\textbf{24.02}^{***}$ \\
vie & 0.75 & 0.93 & $\textbf{24.81}^{***}$ & 0.95 & 1.26 & $\textbf{32.73}^{***}$ \\
zho & 0.77 & 0.84 & $\textbf{9.61}^{***}$ & 1.05 & 1.15 & $\textbf{9.45}^{***}$ \\
\bottomrule
\end{tabular}
\caption{ Mean Global CV (Coefficient of Variation) and Local CV across languages and conditions for captions in \textsc{GROUND-XM3600}.}
\label{tab:xcross_coeffvar}
\end{table}
        \begin{table}[t]
\centering
\setlength{\tabcolsep}{4pt}
\begin{tabular}{l|llll}
\toprule
\textbf{Lang} & \U & \Pp & \D & $[\text{\Pp} + \text{\D}]$ \\
\midrule
ben & 1.05 & 0.99 & 0.94 & 0.93 \\
deu & 1.11 & 1.07 & 1.08 & 1.08 \\
eng & 1.06 & 1.01 & 1.13 & 1.13 \\
fas & 0.87 & 0.83 & 0.93 & 0.91 \\
fra & 1.04 & 1.01 & 1.07 & 1.08 \\
ind & 0.91 & 0.88 & 0.99 & 1.01 \\
kir & 0.83 & 0.83 & 0.87 & 0.88 \\
kor & 0.88 & 0.81 & 0.80 & 0.80 \\
por & 1.01 & 0.96 & 0.99 & 0.99 \\
rus & 0.97 & 0.95 & 0.99 & 1.00 \\
spa & 1.00 & 0.95 & 1.00 & 1.01 \\
tha & 0.91 & 0.83 & 0.86 & 0.86 \\
zho & 0.94 & 0.88 & 1.12 & 1.09 \\
\bottomrule
\end{tabular}
\caption{ Mean Global CV across languages and conditions for paragraph in \textsc{BLOOM-Vist}.}
\label{tab:bloom_coeffvar}
\end{table}

\section{Part-of-Speech Analysis}
    \label{app:pos_analysis}
     \begin{figure*}[t]
    \centering
    \includegraphics[width=\textwidth]{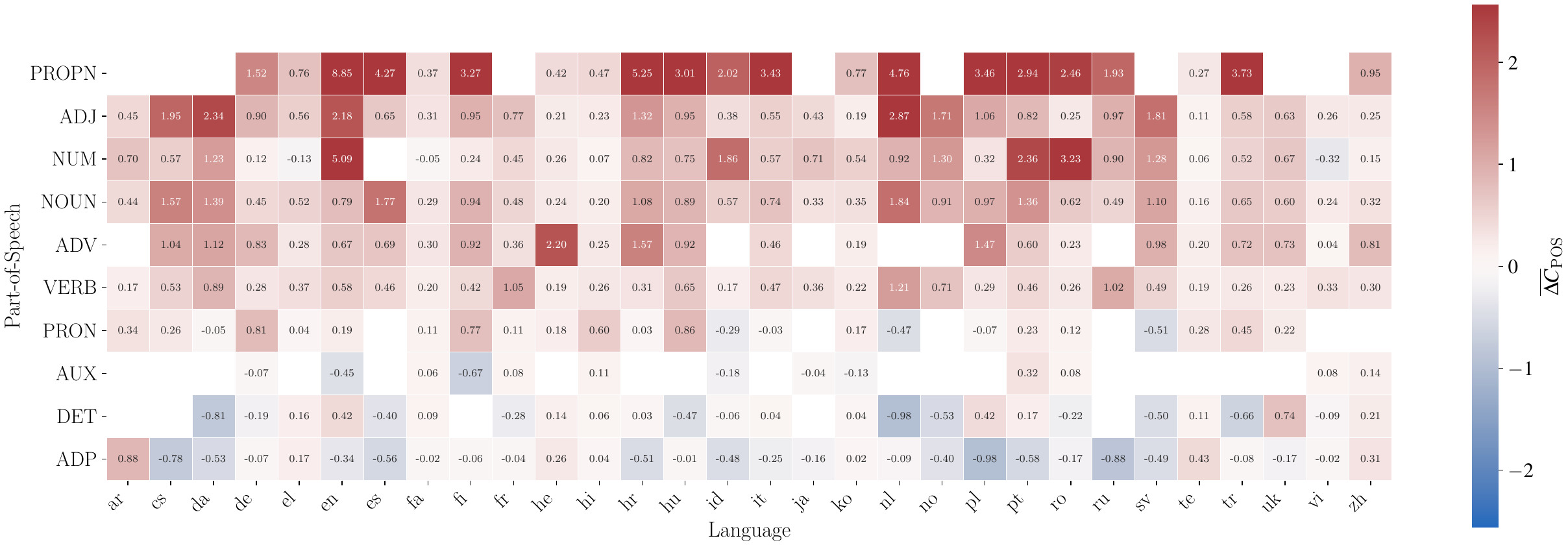}
\caption{Average change in variance contribution ($\overline{\Delta C}_{\text{POS}}$) by POS across languages for UID reduction failure cases in \textsc{Ground-XM3600}. Rows are ordered, from top to bottom, by descending cross-linguistic mean. Intuitively, red cells denote an increased divergence from the sequence mean  under visual grounding (\Pp) relative to text-only (\U) conditions.} 
 \label{fig:XCROSS_heatmap_distribution}
\end{figure*}
     \begin{figure*}[b]
    \centering
    \includegraphics[width=\textwidth]{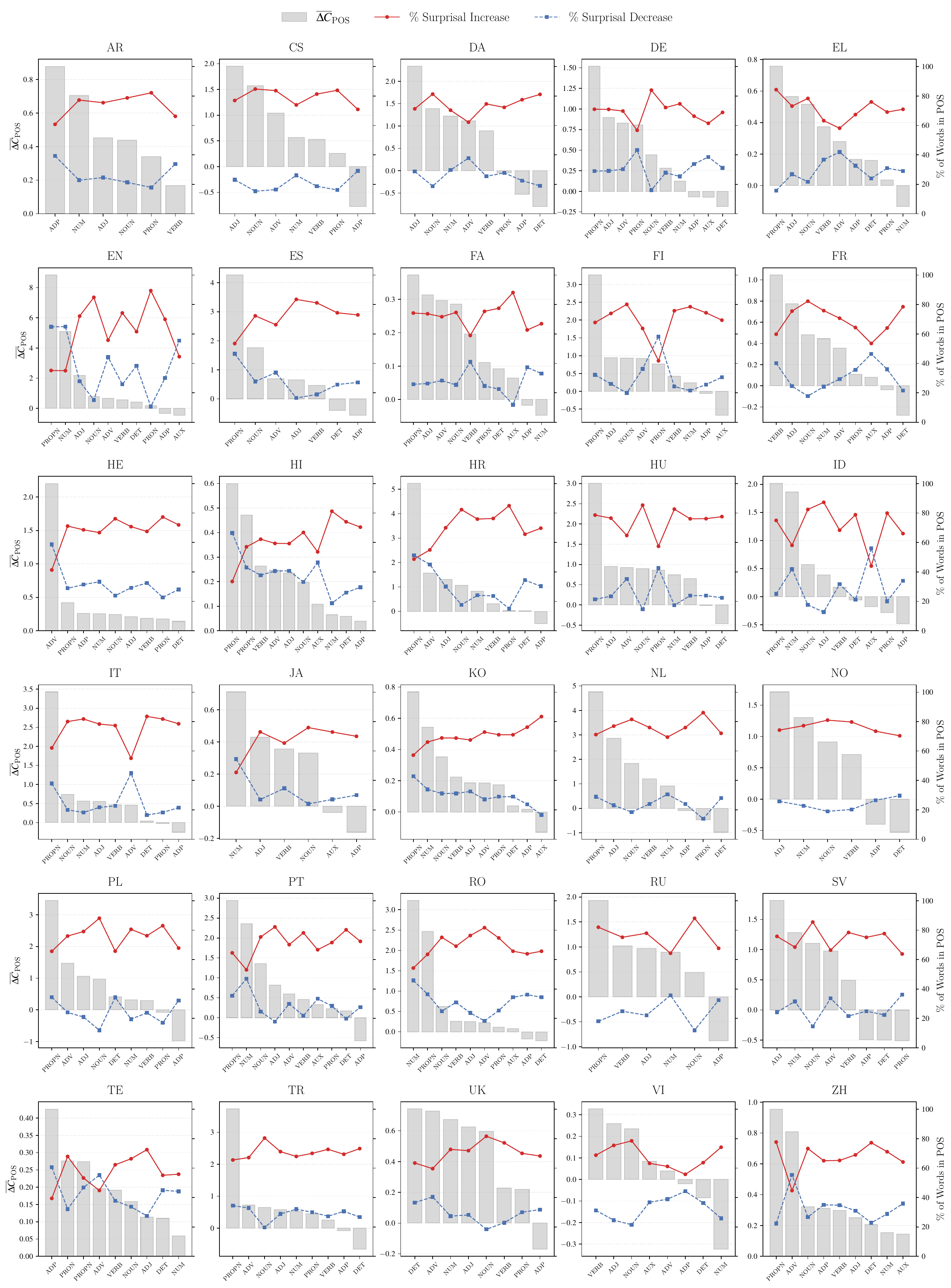}
    \caption{Cross-linguistic dynamics of UID reduction failure across 30 languages in \textsc{Ground-XM3600}. The bar charts represent the average change in variance contribution ($\overline{\Delta C}_{\text{POS}}$) per Part-of-Speech in sentences where visual grounding increases global variance ($\text{UID}_v(\text{\Pp}) > \text{UID}_v(\text{\U})$) (see also \cref{fig:XCROSS_heatmap_distribution}). Line plots illustrate the percentage of words per POS category exhibiting surprisal increases (red circles) versus decreases (blue squares).}
        \label{fig:XCROSS_pos}
\end{figure*}
    The heatmap in \cref{fig:XCROSS_heatmap_distribution}  visualises the average change in variance contribution ($\overline{\Delta C}_{\text{POS}}$, defined in \cref{eq:deltac}), across POS tags and languages, in \textsc{Ground-XM3600} examples. This metric quantifies the extent to which each POS category's contribution to sentence-level variance changes when utterances are grounded on visual perception (\Pp) compared to text-only conditions (\U). 
    The analysis considers only sentences where visual grounding \textit{increased} the global UID value (i.e., where $\text{UID}_v(\Pp) > \text{UID}_v(\U)$). We aggregate results for the 10 most common POS categories in Universal Dependencies (VERB, NOUN, ADJ, ADV, PROPN, ADP, DET, PUNCT, NUM, PRON). A POS-language pair was filtered out if fewer than 50 POS instances per language were present within the identified failure sentences. We define the word-level change in variance contribution  as $
    \Delta C_w = C_{\Pp, w} - C_{\U, w}$
    where $C_{X, w}$ is the contribution of word $w$ to the total sentence variance ($\uidv$) under condition $X \in \{\U, \Pp\}$, i.e.:
    \begin{equation}
    \label{eq:contrib_global}
    C_{X, w} = \frac{(s_{X, w} - \mu_{X})^2}{n}
    \end{equation}
    where $s$ is the word surprisal and $\mu_X$ is the mean surprisal over the $n$ words in the sentence under condition $X$. We then compute the mean of these word-level shifts for each POS group within each language:
    \begin{equation} \label{eq:deltac}
    \overline{\Delta C}_{\text{POS}} = \frac{1}{N_{\text{POS}}} \sum_{w \in \text{POS}} \Delta C_w
    \end{equation}
    where $N_{\text{POS}}$ is the total number of words belonging to a POS category across the identified failure sentences for a given language.
    
    Because this contribution is agnostic to the direction of change in surprisal, we further calculate the percentage of words within each POS category that exhibit an absolute increase ($s_{\mathcal{\Pp},w} > s_{\mathcal{\U},w}$) or decrease ($s_{\mathcal{\Pp},w} < s_{\mathcal{\U},w}$) in surprisal. Results are plotted in \cref{fig:XCROSS_pos}.

\section{Regression Model for RQ3}
    \label{app:BLOOM_ols}

    \paragraph{Design.}
    To model the evolution of surprisal and information density over narrative time across different conditions, we implemented a set of linear mixed-effects regression models with the following structure:
    \begin{equation}
    \begin{split}
    \label{eq:rq3_model}
        y \sim \text{position} \times \text{condition} + \log(\text{length}) \\
        \,+\, (1 + \text{position} \mid \text{story})
    \end{split}
    \end{equation}

    Here, the dependent variable $y$ corresponds to either the mean word-level surprisal or the global UID score ($\uidv$) computed over a given unit---either a sentence or a paragraph. The fixed-effect structure includes the unit’s relative position within the next-higher structural level (i.e., sentence within paragraph or paragraph within story), a categorical variable for context condition (\U, \Pp, \D, \PD), and their interaction. This interaction term allows the slope of surprisal or UID over narrative time to vary across conditions, with the baseline (\U) serving as the reference level. A log-transformed control for unit length (in words) is included to account for length-related variance in surprisal and information distribution.

    The model also includes random intercepts by story to account for between-narrative variability, and random slopes over position by story. This structure tests whether the trajectory of surprisal or UID over discourse position changes systematically as a function of contextual grounding.
    
    \paragraph{Implementation.}
    Models were fit independently for each language and unit type (sentence or paragraph). Sentence-level analyses required that each paragraph contain at least three sentences.
    
    The response variable was regressed on relative position, context, their interaction, and unit length. 
    After fitting, we extracted the fixed-effect slopes per context and dependent variable. Each slope represents the estimated rate of change in surprisal or UID over the course of the narrative, conditional on the availability of contextual information. These results are reported in \cref{tab:BLOOM_mixed_surprisal} and \cref{tab:BLOOM_mixed_uidv}.
        \begin{table*}[t]
\centering
\renewcommand{\arraystretch}{1.2}
\begin{tabular}{c|c|r:r:r:r}
\toprule
\textbf{Lang} & \textbf{Unit} & \U & \Pp & \D & $[\text{\Pp} + \text{\D}]$ \\
\midrule
\multirow[t]{2}{*}{ben} & paragraph & 0.04~n.s & 0.01~n.s & -1.37~*** & -1.19~*** \\
 & sentence & -4.13~*** & -3.27~*** & -0.81~*** & -0.56~*** \\
\cline{1-6}
\multirow[t]{2}{*}{deu} & paragraph & -0.12~n.s & -0.08~n.s & -0.98~*** & -0.71~** \\
 & sentence & -3.25~*** & -2.09~*** & -0.51~*** & -0.31~*** \\
\cline{1-6}
\multirow[t]{2}{*}{eng} & paragraph & 0.24~*** & 0.27~n.s & -1.00~*** & -0.67~*** \\
 & sentence & -2.93~*** & -1.92~*** & -0.64~*** & -0.39~*** \\
\cline{1-6}
\multirow[t]{2}{*}{fas} & paragraph & -0.23~n.s & -0.26~n.s & -3.36~** & -2.84~** \\
 & sentence & -5.83~*** & -3.77~* & -1.13~*** & -0.91~*** \\
\cline{1-6}
\multirow[t]{2}{*}{fra} & paragraph & 0.15~* & 0.17~n.s & -1.00~*** & -0.72~*** \\
 & sentence & -3.48~*** & -2.39~*** & -0.62~*** & -0.41~*** \\
\cline{1-6}
\multirow[t]{2}{*}{ind} & paragraph & 0.19~n.s & 0.35~n.s & -2.15~*** & -1.61~*** \\
 & sentence & -5.18~*** & -4.05~*** & -1.27~*** & -0.90~*** \\
\cline{1-6}
\multirow[t]{2}{*}{kir} & paragraph & 0.25~n.s & 0.28~n.s & -1.76~*** & -1.61~*** \\
 & sentence & -5.11~*** & -4.68~*** & -0.96~*** & -0.76~*** \\
\cline{1-6}
\multirow[t]{2}{*}{kor} & paragraph & -0.17~n.s & -0.11~n.s & -1.62~*** & -1.14~*** \\
 & sentence & -6.15~*** & -3.77~*** & -0.10~*** & 0.13~*** \\
\cline{1-6}
\multirow[t]{2}{*}{por} & paragraph & 0.07~n.s & 0.13~n.s & -0.98~*** & -0.63~*** \\
 & sentence & -4.71~*** & -2.72~*** & -0.08~*** & 0.13~*** \\
\cline{1-6}
\multirow[t]{2}{*}{rus} & paragraph & 0.08~n.s & 0.12~n.s & -1.27~*** & -1.01~*** \\
 & sentence & -3.75~*** & -2.74~*** & -0.79~*** & -0.50~*** \\
\cline{1-6}
\multirow[t]{2}{*}{spa} & paragraph & 0.11~n.s & 0.13~n.s & -1.08~*** & -0.75~*** \\
 & sentence & -3.31~*** & -2.06~*** & -0.36~*** & -0.07~*** \\
\cline{1-6}
\multirow[t]{2}{*}{tha} & paragraph & 0.19~* & 0.35~n.s & -1.53~*** & -1.07~*** \\
 & sentence & -3.42~*** & -2.54~*** & -0.86~*** & -0.62~*** \\
\cline{1-6}
zho & paragraph & 2.23~** & 1.85~n.s & -1.52~*** & -1.10~** \\
\cline{1-6}
\bottomrule
\end{tabular}
\caption{Fixed-effect slope estimates over relative position for \textbf{mean surprisal}, from mixed-effects models (with random effects by story), computed separately per language and discourse unit (sentence, paragraph). Slopes reflect condition-specific trajectories via interaction terms, with \U\ as the baseline. Chinese was excluded from sentence-level analysis due to insufficient longer sentences for reliable estimation. Significance thresholds: $^\ast$ $p<0.05$, $^{\ast\ast}$ $p<0.01$, $^{\ast\ast\ast}$ $p<0.001$, n.s.~(not significant).}
\label{tab:BLOOM_mixed_surprisal}
\end{table*}
        \begin{table*}[t]
\centering
\renewcommand{\arraystretch}{1.2}
\begin{tabular}{c|c|r:r:r:r}
\toprule
\textbf{Lang} & \textbf{Unit} & \U & \Pp & \D & $[\text{\Pp} + \text{\D}]$ \\
\midrule
\multirow[t]{2}{*}{ben} & paragraph & 0.74~n.s & -0.46~n.s & -17.31~*** & -11.90~*** \\
 & sentence & -66.38~*** & -44.07~*** & -9.45~*** & -6.01~*** \\
\cline{1-6}
\multirow[t]{2}{*}{deu} & paragraph & -1.48~n.s & -0.56~n.s & -9.49~*** & -5.66~* \\
 & sentence & -31.17~*** & -16.52~*** & -3.22~*** & -1.27~*** \\
\cline{1-6}
\multirow[t]{2}{*}{eng} & paragraph & 1.97~*** & 2.18~n.s & -8.09~*** & -3.99~*** \\
 & sentence & -25.15~*** & -12.32~*** & -4.40~*** & -1.93~*** \\
\cline{1-6}
\multirow[t]{2}{*}{fas} & paragraph & 9.40~n.s & 6.48~n.s & -25.22~* & -14.72~n.s \\
 & sentence & -54.23~n.s & -19.38~n.s & 2.37~** & -0.79~* \\
\cline{1-6}
\multirow[t]{2}{*}{fra} & paragraph & -0.20~n.s & 0.40~n.s & -7.70~*** & -4.46~*** \\
 & sentence & -28.39~*** & -15.45~*** & -4.18~*** & -2.58~*** \\
\cline{1-6}
\multirow[t]{2}{*}{ind} & paragraph & -2.69~n.s & -0.95~n.s & -19.95~*** & -12.42~*** \\
 & sentence & -47.16~*** & -32.37~*** & -9.67~*** & -6.76~*** \\
\cline{1-6}
\multirow[t]{2}{*}{kir} & paragraph & 1.97~n.s & 1.59~n.s & -19.30~*** & -17.86~*** \\
 & sentence & -55.03~*** & -48.58~*** & -8.16~*** & -6.96~*** \\
\cline{1-6}
\multirow[t]{2}{*}{kor} & paragraph & -5.33~* & -4.52~n.s & -19.99~*** & -10.69~* \\
 & sentence & -88.33~*** & -39.77~*** & -3.17~*** & -0.98~*** \\
\cline{1-6}
\multirow[t]{2}{*}{por} & paragraph & -1.17~n.s & -0.48~n.s & -8.18~*** & -4.10~** \\
 & sentence & -46.08~*** & -22.87~*** & -1.31~*** & -0.18~*** \\
\cline{1-6}
\multirow[t]{2}{*}{rus} & paragraph & -0.74~n.s & -0.39~n.s & -11.81~*** & -8.38~*** \\
 & sentence & -35.96~*** & -23.21~*** & -7.11~*** & -4.54~*** \\
\cline{1-6}
\multirow[t]{2}{*}{spa} & paragraph & 0.72~n.s & 0.99~n.s & -9.03~*** & -4.75~*** \\
 & sentence & -31.38~*** & -14.77~*** & -2.81~*** & -0.27~*** \\
\cline{1-6}
\multirow[t]{2}{*}{tha} & paragraph & -0.72~n.s & 0.49~n.s & -15.20~*** & -7.15~*** \\
 & sentence & -38.14~*** & -20.21~*** & -7.07~*** & -4.45~*** \\
\cline{1-6}
zho & paragraph & 15.42~n.s & 13.83~n.s & -27.87~** & -16.64~n.s \\
\cline{1-6}
\bottomrule
\end{tabular}
\caption{Fixed-effect slope estimates over relative position for $\bm{\mathrm{UID}}_{v}$, from mixed-effects models (with random effects by story), computed separately per language and discourse unit (sentence, paragraph). Slopes reflect condition-specific trajectories via interaction terms, with \U\ as the baseline. Chinese was excluded from sentence-level analysis due to insufficient longer sentences for reliable estimation. Significance thresholds: $^\ast$ $p<0.05$, $^{\ast\ast}$ $p<0.01$, $^{\ast\ast\ast}$ $p<0.001$, n.s.~(not significant).}

\label{tab:BLOOM_mixed_uidv}
\end{table*}
    \section{Surprisal Reduction Densities}
\label{app:density_plot}
    Figures~8--20 display the distribution of positive surprisal reductions across the relative position of words within sentences and paragraphs, for each language in \textsc{BloomVist}. The densities are estimated by computing histograms over normalized positions and then smoothed via Gaussian filters. The three curves in each plot correspond to:
    
    \begin{itemize}
      \item $\DeltaP$: reduction due to local visual context (only previous image) vs.\ uncontextualised text, (i.e., $\text{Surprisal}(\text{\U} - \text{\Pp})$).
      \item $\DeltaD$: reduction due to discourse-level context vs.\ sentence-level context (i.e., $\text{Surprisal}(\text{\U} - \text{\D})$).\looseness-1
      \item $\DeltaPD$: reduction due to global visual context  vs.\ text-only discourse level (i.e., $\text{Surprisal}(\text{\D} - [\text{\Pp} + \text{\D}]$)).
    \end{itemize}

\section{Surprisal Discontinuities at Discourse Boundaries}
    The tables below report differences around sentence (\cref{tab:directional_window_deltas_sentence}) and paragraph (\cref{tab:directional_window_deltas_paragraph}) boundaries for each language in \textsc{BloomVIST}, quantifying information spikes at discourse unit onsets for varying window sizes.

\section{Compute Budget}
    Surprisal values from \textsc{BloomVIST} were extracted using the 4B-parameter Gemma 3 model, loaded with HuggingFace's Transformers library.\footnote{\url{https://pypi.org/project/transformers/}} The full extraction required approximately 24 hours of compute on a single \textsc{NVIDIA A100-80GB GPU}.

\onecolumn

\newcommand{\includebloomplot}[2]{%
  \begin{figure}[ht]
    \centering
    \includegraphics[width=\textwidth]{#1}
    \caption{Surprisal reduction density for \texttt{#2}.}
    \label{fig:#2-density}
  \end{figure}
}

\setlength{\intextsep}{2pt}   
\setlength{\textfloatsep}{2pt}
\setlength{\floatsep}{2pt}

\includebloomplot{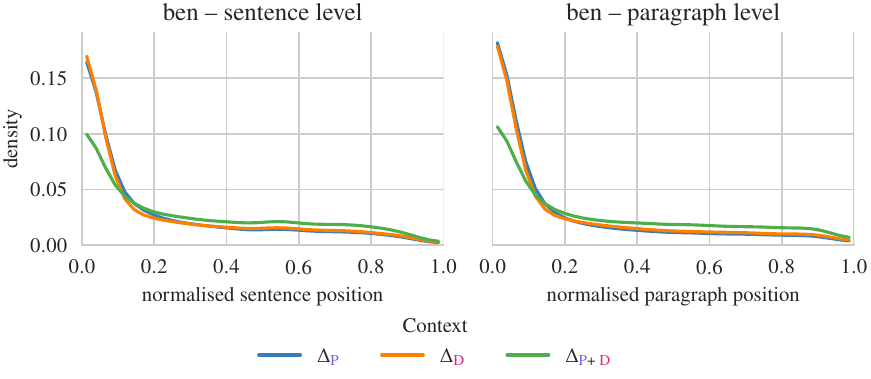}{ben}
\includebloomplot{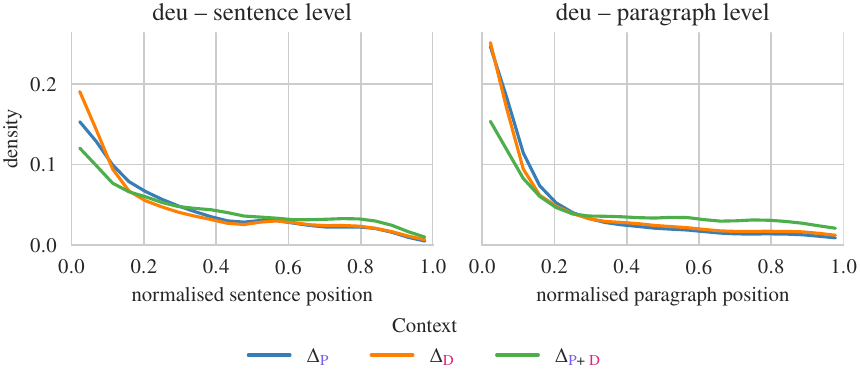}{deu}
\includebloomplot{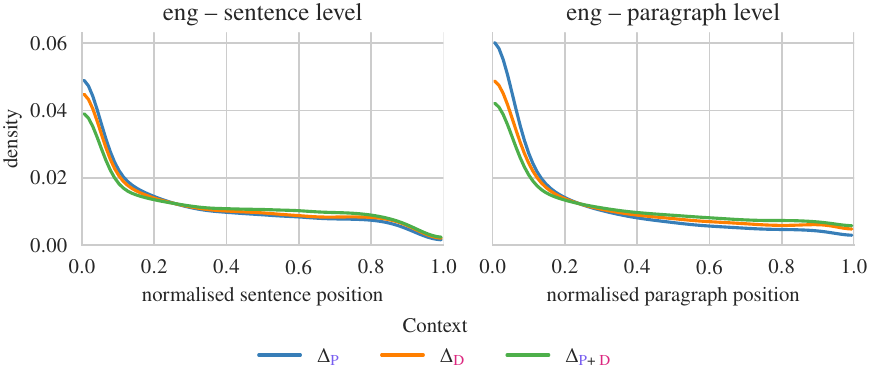}{eng}
\includebloomplot{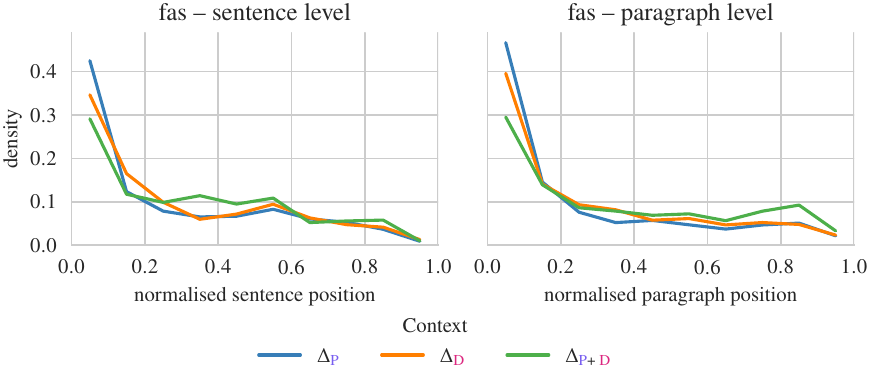}{fas}
\includebloomplot{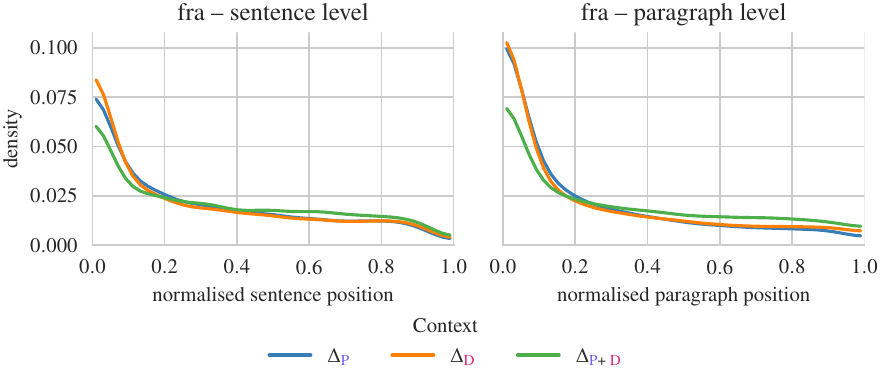}{fra}
\includebloomplot{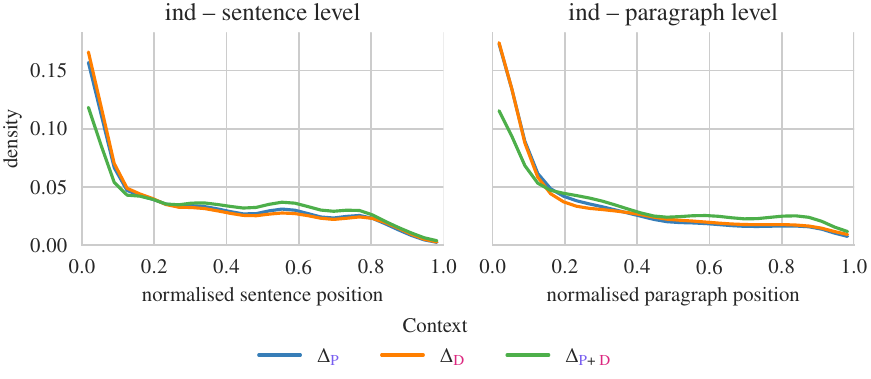}{ind}
\includebloomplot{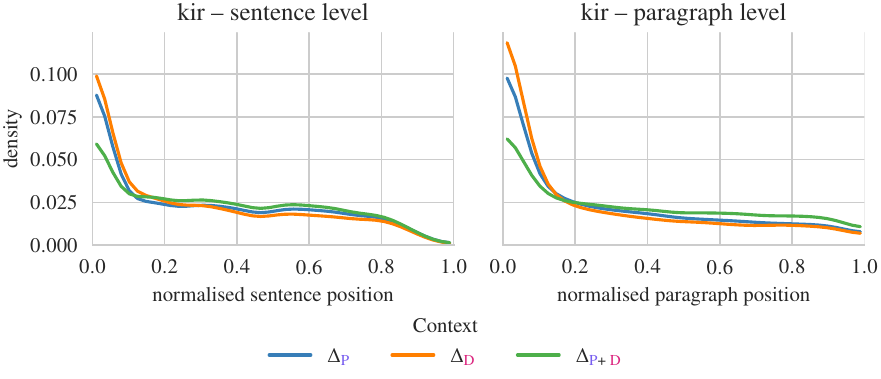}{kir}
\includebloomplot{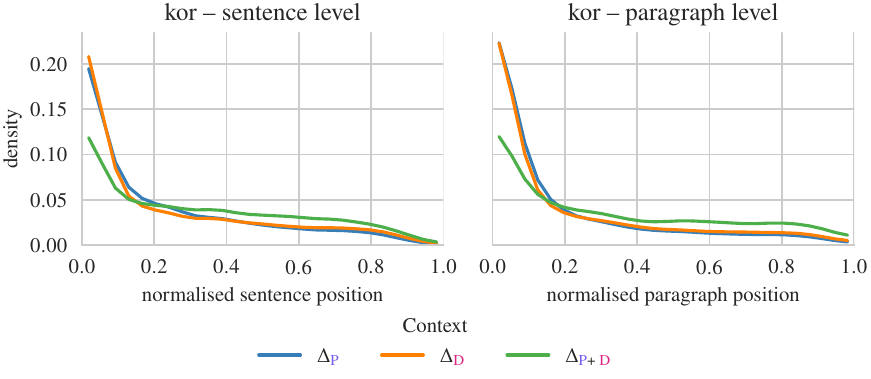}{kor}
\includebloomplot{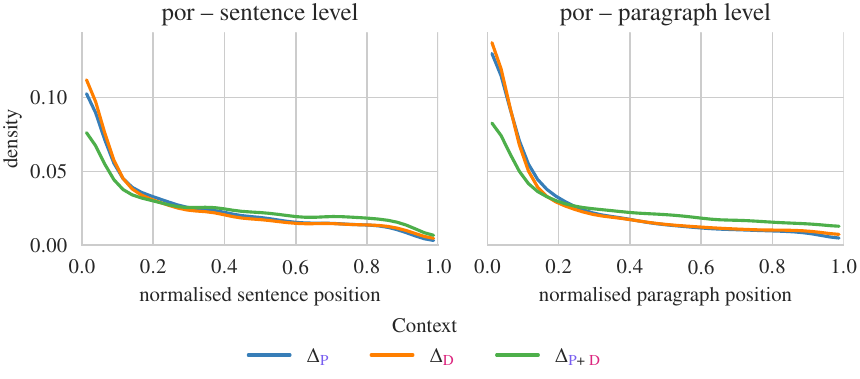}{por}
\includebloomplot{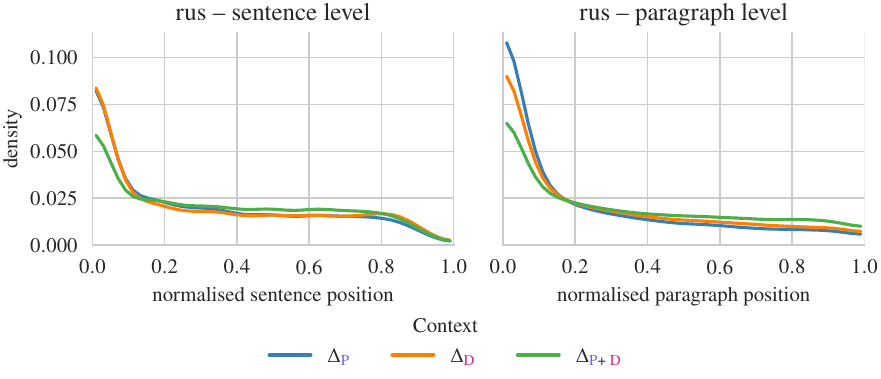}{rus}
\includebloomplot{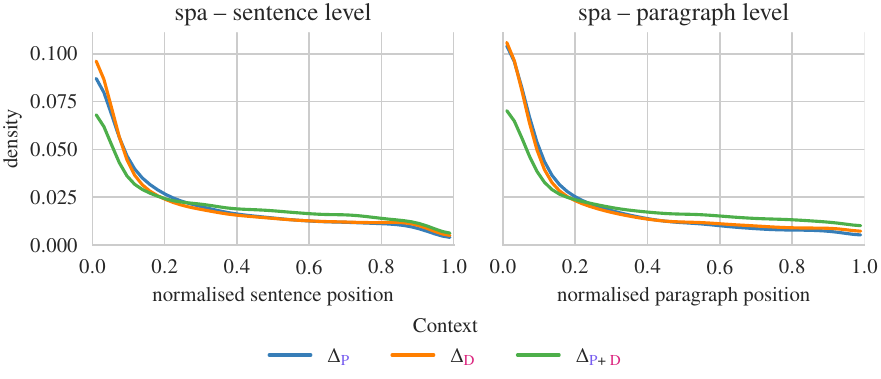}{spa}
\includebloomplot{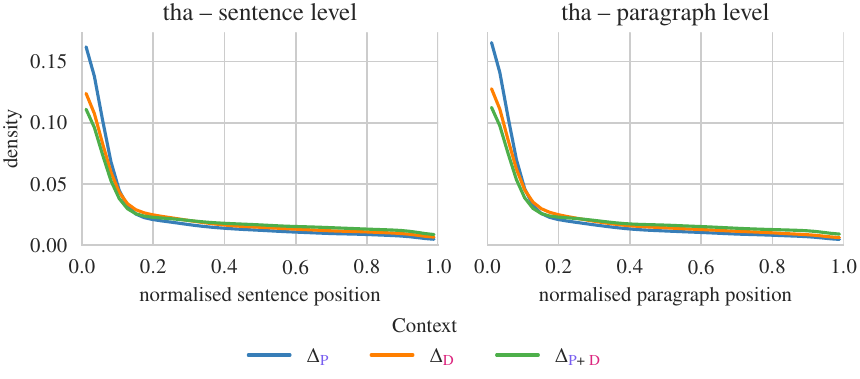}{tha}
\includebloomplot{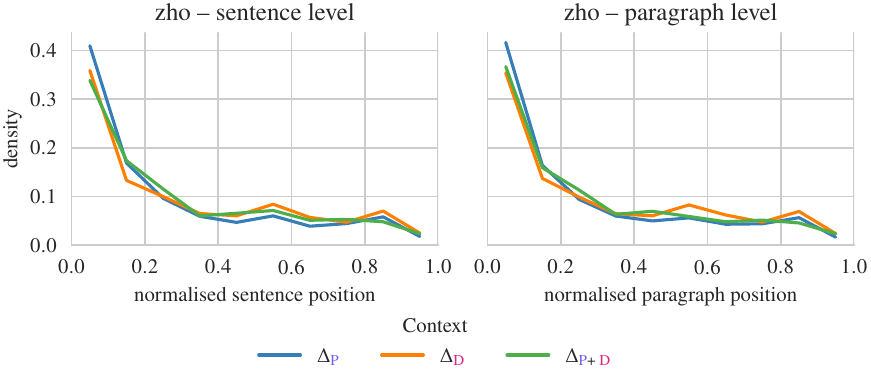}{zho}

\twocolumn

    \begin{table*}[t]
\centering
\setlength{\tabcolsep}{6pt}
\renewcommand{\arraystretch}{1.0}
\begin{tabular}{c|c|r:r:r:r}
\toprule
\textbf{Lang} & $\Delta_w$ & \U & \Pp & \D & $[\text{\Pp} + \text{\D}]$ \\
\midrule
\multirow{3}{*}{ben} & $\Delta_{1}$ & -33.798 & -27.867 & -7.777 & -7.036 \\
& $\Delta_{2}$ & -21.496 & -17.803 & -5.861 & -5.360 \\
& $\Delta_{3}$ & -15.853 & -13.089 & -4.450 & -3.980 \\
\cline{1-6}
\multirow{3}{*}{deu} & $\Delta_{1}$ & -18.408 & -14.799 & -4.418 & -4.002 \\
& $\Delta_{2}$ & -15.674 & -12.237 & -3.351 & -3.141 \\
& $\Delta_{3}$ & -12.439 & -9.234 & -2.398 & -2.246 \\
\cline{1-6}
\multirow{3}{*}{eng} & $\Delta_{1}$ & -13.444 & -9.272 & -4.644 & -3.685 \\
& $\Delta_{2}$ & -10.643 & -7.339 & -3.264 & -2.682 \\
& $\Delta_{3}$ & -8.773 & -6.126 & -2.540 & -2.100 \\
\cline{1-6}
\multirow{3}{*}{fas} & $\Delta_{1}$ & -28.405 & -18.189 & -6.006 & -5.242 \\
& $\Delta_{2}$ & -18.923 & -13.385 & -3.971 & -3.856 \\
& $\Delta_{3}$ & -14.230 & -10.333 & -3.066 & -2.908 \\
\cline{1-6}
\multirow{3}{*}{fra} & $\Delta_{1}$ & -17.818 & -13.930 & -4.165 & -3.513 \\
& $\Delta_{2}$ & -13.930 & -10.357 & -2.882 & -2.447 \\
& $\Delta_{3}$ & -11.239 & -8.360 & -2.382 & -2.051 \\
\cline{1-6}
\multirow{3}{*}{ind} & $\Delta_{1}$ & -21.821 & -17.403 & -4.656 & -3.873 \\
& $\Delta_{2}$ & -15.581 & -12.719 & -3.133 & -2.731 \\
& $\Delta_{3}$ & -11.727 & -9.542 & -2.233 & -1.909 \\
\cline{1-6}
\multirow{3}{*}{kir} & $\Delta_{1}$ & -28.612 & -25.267 & -7.203 & -6.643 \\
& $\Delta_{2}$ & -20.486 & -18.859 & -4.737 & -4.297 \\
& $\Delta_{3}$ & -15.758 & -14.631 & -3.302 & -2.943 \\
\cline{1-6}
\multirow{3}{*}{kor} & $\Delta_{1}$ & -31.875 & -23.544 & -6.223 & -5.794 \\
& $\Delta_{2}$ & -20.749 & -14.585 & -3.998 & -3.802 \\
& $\Delta_{3}$ & -15.638 & -10.827 & -2.874 & -2.706 \\
\cline{1-6}
\multirow{3}{*}{por} & $\Delta_{1}$ & -17.604 & -12.163 & -3.229 & -2.869 \\
& $\Delta_{2}$ & -14.695 & -10.185 & -2.405 & -2.222 \\
& $\Delta_{3}$ & -11.453 & -7.802 & -1.827 & -1.693 \\
\cline{1-6}
\multirow{3}{*}{rus} & $\Delta_{1}$ & -21.067 & -15.896 & -5.535 & -4.513 \\
& $\Delta_{2}$ & -15.306 & -11.684 & -4.033 & -3.392 \\
& $\Delta_{3}$ & -11.230 & -8.438 & -2.552 & -2.086 \\
\cline{1-6}
\multirow{3}{*}{spa} & $\Delta_{1}$ & -17.153 & -12.912 & -4.245 & -3.693 \\
& $\Delta_{2}$ & -12.860 & -9.129 & -2.745 & -2.349 \\
& $\Delta_{3}$ & -10.326 & -7.234 & -2.128 & -1.808 \\
\cline{1-6}
\multirow{3}{*}{tha} & $\Delta_{1}$ & -25.718 & -16.269 & -4.808 & -2.878 \\
& $\Delta_{2}$ & -15.599 & -10.085 & -2.472 & -1.338 \\
& $\Delta_{3}$ & -11.468 & -7.542 & -1.626 & -0.809 \\
\cline{1-6}
\multirow{3}{*}{zho} & $\Delta_{1}$ & -29.135 & -20.430 & -6.761 & -4.975 \\
& $\Delta_{2}$ & -18.282 & -12.742 & -4.687 & -3.503 \\
& $\Delta_{3}$ & -13.019 & -9.277 & -3.033 & -2.332 \\
\cline{1-6}
\bottomrule
\end{tabular}
\caption{Mean surprisal difference across \textbf{sentence} boundaries ($\Delta_w$) for window sizes $w = 1, 2, 3$, reported per language. For each sentence transition, we subtract the average surprisal of the final $w$ words of the preceding sentence from that of the first $w$ words of the following one. More negative values indicate sharper surprisal spikes at sentence onsets, reflecting stronger deviations from uniform information flow across sentence boundaries.}
\label{tab:directional_window_deltas_sentence}
\end{table*}
    \begin{table*}[t]
\centering
\setlength{\tabcolsep}{6pt}
\renewcommand{\arraystretch}{1.0}
\begin{tabular}{c|c|r:r:r:r}
\toprule
\textbf{Lang} & $\Delta_w$ & \U & \Pp & \D & $[\text{\Pp} + \text{\D}]$ \\
\midrule
\multirow{3}{*}{ben} & $\Delta_{1}$ & -33.927 & -27.903 & -7.909 & -7.095 \\
& $\Delta_{2}$ & -21.433 & -17.691 & -5.919 & -5.350 \\
& $\Delta_{3}$ & -15.800 & -12.983 & -4.470 & -3.947 \\
\cline{1-6}
\multirow{3}{*}{deu} & $\Delta_{1}$ & -18.890 & -15.328 & -4.508 & -4.064 \\
& $\Delta_{2}$ & -15.885 & -12.417 & -3.393 & -3.175 \\
& $\Delta_{3}$ & -12.516 & -9.246 & -2.368 & -2.196 \\
\cline{1-6}
\multirow{3}{*}{eng} & $\Delta_{1}$ & -12.443 & -8.348 & -3.913 & -3.196 \\
& $\Delta_{2}$ & -9.996 & -6.667 & -2.747 & -2.319 \\
& $\Delta_{3}$ & -8.264 & -5.576 & -2.108 & -1.784 \\
\cline{1-6}
\multirow{3}{*}{fas} & $\Delta_{1}$ & -27.202 & -16.985 & -5.372 & -4.548 \\
& $\Delta_{2}$ & -17.732 & -12.085 & -3.205 & -3.021 \\
& $\Delta_{3}$ & -13.460 & -9.521 & -2.597 & -2.377 \\
\cline{1-6}
\multirow{3}{*}{fra} & $\Delta_{1}$ & -17.932 & -14.049 & -4.070 & -3.502 \\
& $\Delta_{2}$ & -13.919 & -10.280 & -2.755 & -2.402 \\
& $\Delta_{3}$ & -11.134 & -8.186 & -2.225 & -1.964 \\
\cline{1-6}
\multirow{3}{*}{ind} & $\Delta_{1}$ & -22.326 & -17.783 & -4.632 & -3.841 \\
& $\Delta_{2}$ & -15.722 & -12.798 & -3.028 & -2.622 \\
& $\Delta_{3}$ & -11.914 & -9.668 & -2.218 & -1.885 \\
\cline{1-6}
\multirow{3}{*}{kir} & $\Delta_{1}$ & -28.969 & -25.674 & -7.307 & -6.724 \\
& $\Delta_{2}$ & -20.228 & -18.679 & -4.742 & -4.294 \\
& $\Delta_{3}$ & -15.590 & -14.501 & -3.328 & -2.948 \\
\cline{1-6}
\multirow{3}{*}{kor} & $\Delta_{1}$ & -31.906 & -23.686 & -6.366 & -5.906 \\
& $\Delta_{2}$ & -20.678 & -14.559 & -4.066 & -3.850 \\
& $\Delta_{3}$ & -15.589 & -10.820 & -2.957 & -2.768 \\
\cline{1-6}
\multirow{3}{*}{por} & $\Delta_{1}$ & -17.340 & -12.061 & -3.323 & -2.979 \\
& $\Delta_{2}$ & -14.404 & -10.039 & -2.436 & -2.259 \\
& $\Delta_{3}$ & -11.305 & -7.743 & -1.894 & -1.764 \\
\cline{1-6}
\multirow{3}{*}{rus} & $\Delta_{1}$ & -21.450 & -16.293 & -5.547 & -4.662 \\
& $\Delta_{2}$ & -15.427 & -11.786 & -3.981 & -3.434 \\
& $\Delta_{3}$ & -11.436 & -8.606 & -2.620 & -2.210 \\
\cline{1-6}
\multirow{3}{*}{spa} & $\Delta_{1}$ & -17.171 & -12.787 & -4.072 & -3.586 \\
& $\Delta_{2}$ & -12.954 & -9.054 & -2.680 & -2.324 \\
& $\Delta_{3}$ & -10.404 & -7.173 & -2.106 & -1.820 \\
\cline{1-6}
\multirow{3}{*}{tha} & $\Delta_{1}$ & -25.826 & -16.347 & -4.758 & -2.835 \\
& $\Delta_{2}$ & -15.534 & -10.017 & -2.423 & -1.281 \\
& $\Delta_{3}$ & -11.407 & -7.479 & -1.616 & -0.785 \\
\cline{1-6}
\multirow{3}{*}{zho} & $\Delta_{1}$ & -28.585 & -20.354 & -6.126 & -4.386 \\
& $\Delta_{2}$ & -17.784 & -12.421 & -4.135 & -2.950 \\
& $\Delta_{3}$ & -12.591 & -8.992 & -2.589 & -1.904 \\
\cline{1-6}
\bottomrule
\end{tabular}
\caption{Mean surprisal difference across \textbf{paragraph} boundaries ($\Delta_w$) for window sizes $w = 1, 2, 3$, reported per language. For each transition, we subtract the average surprisal of the final $w$ words of a paragraph from that of the first $w$ words of the following one. More negative values indicate larger spikes in surprisal at paragraph onsets, reflecting greater non-uniformity in information flow across discourse boundaries.}
\label{tab:directional_window_deltas_paragraph}
\end{table*}
    \newpage

\end{document}